\pdfoutput=1

\documentclass{article}
\usepackage[utf8]{inputenc} % allow utf-8 input
\usepackage[T1]{fontenc}    % use 8-bit T1 fonts
\usepackage[dvipsnames,table]{xcolor}
\usepackage{graphicx}
\definecolor{citecol}{HTML}{6F130C}
\definecolor{tableofcontent}{HTML}{1F4A83}
\definecolor{urlcol}{HTML}{2470D8}
\usepackage{hyperref}       % hyperlinks
\usepackage{booktabs}       % professional-quality tables
\usepackage{amsfonts}       % blackboard math symbols
\usepackage{nicefrac}       % compact symbols for 1/2, etc.
\usepackage{microtype}      % microtypography
\usepackage{subfigure}
\usepackage{amsthm}
\usepackage{amssymb}
\newtheorem{lem}{Lemma}
\theoremstyle{remark}
\newtheorem{remark}{Remark}
\usepackage{amsmath}
\usepackage{algorithm}
\usepackage{algorithmic}
\usepackage{multirow}
\usepackage{enumitem}
\usepackage{callouts}
\usepackage{amsthm}
\usepackage{comment}
\usepackage{natbib}
\DeclareMathOperator*{\argmin}{arg\,min}
\theoremstyle{definition}
\newtheorem{definition}{Definition}[section]
\newtheorem{theorem}{Theorem}[section]

\newcounter{ToDo}
\newcounter{gaocomm}
\newcounter{wangcomm}
\newcounter{Note}
\definecolor{blue-violet}{rgb}{0.00,0.75,0.90}
\definecolor{mygreen}{rgb}{0.0, 0.5, 0.0}
\definecolor{awesome}{rgb}{1.0, 0.13, 0.32}
\definecolor{bostonuniversityred}{rgb}{1.0, 0.0, 0.0}
\newcommand{\EScomment}[1]{{\color{black} #1}}

\title{Generalized Laplacian Regularized Framelet Graph Neural Networks}

\author{Zhiqi Shao\footnote{University of Sydney 
   (\texttt{zhiqi.shao@sydney.edu.au}, \texttt{andi.han@uni.sydney.edu.au}, 
   \texttt{andrey.vasnev@uni.sydney.edu.au}, 
   \texttt{junbin.gao@sydney.edu.au})}
\and Andi Han\footnotemark[1]
\and Dai Shi\footnote{Western Sydney University \texttt{(dai.shi@sydney.edu.au})\\
\text{Zhiqi Shao, Andi Han and Dai Shi are with equal contribution.}}\footnotemark[1]
\and Andrey Vasnev \footnotemark[1]
\and Junbin Gao\footnotemark[1]
}

\begin{document}
\maketitle
\begin{abstract}
Graph neural networks (GNNs) have achieved remarkable results for various graph learning tasks. However, one of the recent challenges for GNNs is to adapt to different types of graph inputs, such as heterophilic graph datasets in which linked nodes are more likely to contain a different class of labels and features. Accordingly, an ideal GNN model should adaptively accommodate all types of graph datasets with different labeling distributions. In this paper, we tackle this challenge by proposing a regularization framework on graph framelet with the regularizer induced from graph $p$-Laplacian. By adjusting the value of $p$, the $p$-Laplacian based regularizer restricts the solution space of graph framelet into the desirable region based on the graph homophilic features. We propose an algorithm to effectively solve a more generalized regularization problem and prove that the algorithm imposes a ($p$-Laplacian based) spectral convolution and diagonal scaling operation to the framelet filtered node features. % double-filtering effect on the graph inputs.
Furthermore, we analyze the denoising power of the proposed model and compare it with the predefined framelet denoising regularizer. Finally, we conduct empirical studies to show the prediction power of the proposed model in both homophily undirect and heterophily direct graphs with and without noises. Our proposed model shows significant improvements compared to multiple baselines, and this suggests the effectiveness of combining graph framelet and $p$-Laplacian.
%\keywords{Graph Neural Networks \and Graph Framelets \and $p$-Laplacian Regularization } 
\end{abstract}

%\linenumbers 

\section*{Highlight}
% \EScomment{Try to summarize our contribution here, leave that to you }
\begin{itemize}
    \item We define two types of new GNN layers by introducing $p$-Laplacian regularizers to both decomposed and reconstructed framelets.
    
    \item We provide an iterative algorithm to fit the proposed models and explore an alternative algorithm developed from the F-norm Locality Preserving Projection (LPP).

    \item We prove that our algorithm  provides a sequence of spectral graph convolutions and diagonal scaling over framelet-filtered graph signals and this further explains the interaction between $p$-Laplacian regularizer and framelet.
    
    \item We link the proposed $p$-Laplacian regularizer with the previous work of framelet regularizer to illustrate the denoising power of our model.   
    % To demonstrate the denoising power, we build a connection between the proposed $p$-Laplacian regularizer and the previous work of framelet regularizer.
    \item We test the model on homophilic (undirected) and heterophilic (directed) graphs. To our best knowledge, we are the first to explore the possibility of applying the framelet GCN for directed graphs
\end{itemize}



\section{Introduction}

% \Question{1. However, other works such as PGNN also have such ability. What are the advantages of the proposed models over PGNN in adapting to different types of networks? The authors are strongly encouraged to clarify this point since this is the main motivation.}  {\color{red}You comment looks okay.}

% \question{2. the space and time complexity of the proposed optimization algorithm, one reviewer questioned about the complexity of generating $\mathcal W$ from 11 to 13. } \AHcomment{The time and space complexity of computing the framelet transform is $O(N^2(LJ+1)d)$.}

% \question{It seems this is the correct two line format for submission, please help to adjust equation size}

% \question{3. The format file can not process  | | the blocking sign. this is why we have 16 errors } \AHcomment{Use $\vert$ instead}
% \question{I excluded the heading of the page, which was set as SpringerNature2021Template}

Graph neural networks (GNNs) have demonstrated remarkable ability for graph learning tasks \citep{BronsteinBrunaLeCunSzlamVandergheynst2017,WuPanChenLongZhangYu2020,ZhangCuiZhu2020,ZhouCuiHuZhangYangLiuWangLiSun2020}. The input to GNNs is the so-called graph data which records useful features and structural information among data. Such data are widely seen in many fields, such as biomedical science \citep{Ahmedt-AristizabalArminDenmanFookesPetersson2021}, social networks \citep{FanMaLiHeZhaoTangYin2019}, and recommend systems \citep{WuSunZhangXieCui2020}. GNN models can be broadly categorized into spectral and spatial methods. The spatial methods such as MPNN \citep{gilmer2017neural}, GAT \citep{velivckovic2017graph} and GIN \citep{xu2018powerful} utilize the message passing mechanism to propagate node feature information based on their neighbours \citep{ScarselliGoriTsoiHagenbuchnerMonfardini2009}. On the other hand, the spectral methods including ChebyNet \citep{defferrard2016convolutional}, GCN \citep{KipfWelling2016} and BernNet \citep{he2021bernnet} are derived from the classic convolutional networks, treating the input graph data as signals (i.e., a function with the domain of graph nodes) \citep{OrtegaFrossardKovacevicMouraVandergheynst2018}, and filtering signals in the Fourier domain \citep{BrunaZarembaSzlamLeCun2013,defferrard2016convolutional}. Among various spectral-based GNNs, the graph framelets GCN \citep{zheng2022decimated}, which is constructed on a type of wavelet frame analogized from the identical concept defined on manifold \citep{Dong2017}, further separates the input signal by predefined low-pass and high-pass filtering functions and resulting a convolution-type model as in the studies \citep{chendirichlet,ZhengZhouGaoWangLioLiMontufar2021,zheng2022decimated}. The graph framelet shows great flexibility in terms of controlling both low and high-frequency information with robustness to noise and thus in general possesses high prediction power in multiple graph learning tasks \citep{chendirichlet,yang2022quasi,ZhengZhouGaoWangLioLiMontufar2021,ZhouLiZhengWangGao2021,zhou2021spectral,ZouHanLinGao2022}.

Along with the path of developing more advanced models, one of the major challenges in GNN is identified from the aspect of data consistency. For example, many GNN models \citep{xu2018powerful,WuPanChenLongZhangYu2020,zhu2020graph,WuSunSunSun2021,LiuLiLiZhou2022} are designed based on the homophily assumption i.e., nodes with similar features or labels are often linked with each other. Such phenomenon can be commonly observed in citation networks \citep{ciotti2016homophily} which have been widely used as benchmarks in GNNs empirical studies. However, the homophily assumption may not always hold, and its opposite, i.e. Heterophily, can be observed quite often in many real-world datasets in which the linked nodes are more likely to contain different class labels and features \citep{zheng2022graph}. For example, in online transaction networks, fraudsters are more likely to link to customers instead of other fraudsters \citep{pandit2007netprobe}. The GNNs designed under homophilic assumption are deemed unsuitable for heterophilic graphs. It is evident from their significant performance degradation \citep{zheng2022graph}.  The reason is that the class of heterophilic graphs contains heterogeneous instances and hence the signals should be sharpened rather than smoothed out. An ideal framework for learning on graphs should be able to accommodate both homophilic and heterophilic scenarios.

\EScomment{One of the active aspects to resolve the GNN adaption challenge is by regularizing the solution of GNNs via the perspective of optimization. The work done by \cite{ZhuWangShiJiCui2021} unified most of state-of-the-art GNNs as an optimization framework.} Furthermore, one of the recent works \citep{FuZhaoBian2022} considered assigning an adjustable $ p$-Laplacian regularizer to the (discrete) graph regularization problem that is conventionally treated as a way of producing GNN outcomes (i.e., Laplacian smoothing). In view of the fact that the classic graph Laplacian regularizer measures the graph signal energy along edges under  $L_2$ metric, it would be beneficial if GNN could be adapted to heterophilic graphs under $L_p$ metric ($1 \leq p < 2$). Given that $L_1$ metric is more robust to high-frequency signals, a higher model discriminative power is preserved. The early work \citep{HuSunGaoHuYin2018} has demonstrated the advantage of adopting $L_1$ metric in the Locality Preserving Projection (LPP) model. In addition, the recently proposed $p$-Laplacian GNN \citep{FuZhaoBian2022} adaptively modifies aggregation weights by exploiting the variance of node embeddings on edges measured by the graph gradient. % this method was initial adopted by \citep{ZhouSchoelkopf2005}. 
With a further investigation on the application of $p$-Laplacian, \cite{LuoHuangDingNie2010} suggested an efficient gradient descent optimization strategy to construct the $p$-Laplacian embedding space that guarantees convergence to viable local solutions. \EScomment{Although the $p$-Laplacian based optimization scheme has been successfully lodged in basic GNN model \citep{FuZhaoBian2022}, whether it can be further incorporated with more advanced GNN model (i.e., graph framelets) with higher prediction power and flexibility is still unclear. Specifically, one may be interested in identifying whether the advantage of deploying $p$-Laplacian in GNN still remains in graph framelet or what is the exact functionality of $p$-Laplacian interacting with the feature representations generated both low and high pass framelet domains. In addition, as graph framelets have shown great potential in terms of the graph noise reduction \citep{zheng2022decimated,yang2022quasi,ZouHanLinGao2022}, whether the inclusion of $p$-Laplacian regularizer could enhance/dilute such power remains unclear. These research gaps inspire us to incorporate $p$-Laplacian into graph framelets and explore further for the underlying relationship between them.}

% Since the $p$-Laplacian regularized optimization problem has no closed-form solution except for the case when $p=2$  \citep{ZhouSchoelkopf2005}. Thus an iterative algorithm (refer to Eq. \eqref{Eq:iter} for more details) is proposed to approximate the solution and each iteration can be interpreted as a GNN layer \citep{ZhouLiZhengWangGao2021}. 

\EScomment{Since the $p$-Laplacian regularized optimization problem lacks a closed-form solution, except for when $p = 2$ \citep{ZhouSchoelkopf2005}, an iterative algorithm is suggested to estimate the solution and each iteration can be analogized to a GNN layer \citep{ZhouLiZhengWangGao2021}).} The solution of such an algorithm is defined by the first-order condition of the optimization problem. As a result, one can relate this to the implicit layer approach \citep{WangWangYangLin2022,ZhangGuMateusz2020} which has the potential of avoiding over-smoothing issues since the adjusted node feature will be re-injected in each iteration step. By adjusting the value of $p$, both $p$-Laplacian GNN and the algorithm are capable of producing  learning outcomes with different discriminative levels and thus able to handle both homophilic and heterophilic graphs. Similar work has been done by \cite{frecon2022bregman} which presents a framework based on the Bregman layer to fulfill the bi-level optimization formulations.
% Similarly, \citep{frecon2022bregman}  Furthermore, the $p$-Laplacian regularized optimization offered a framework that capable of handling both heterophilc and homophilic graphs with an adjustable $p$\citep{FuZhaoBian2022}.
% \question{This is where the editors questioning about, it seems we did not make our motivation very clear, and they ask the reason of combination, I think we can not say to reap the benefit of $p$-lap.}
To reap the benefit of $p$-Laplacian regularization in the framelet domain, in this paper, we propose two $p$-Laplacian Regularized Framelet GCN models which are named $p$-Laplacian Undecimated Framelet GCN (pL-UFG) and $p$-Laplacian Fourier Undecimated Framelet GCN (pL-fUFG). In these models, the $p$-Laplacian regularization can be applied on either the framelet domain or the well-known framelet Fourier domain. The $ p$-Laplacian-based regularizer incurs a penalty to both low and high-frequency domains created by framelet, producing flexible models that are capable of adapting different types of graph inputs (i.e., homophily and heterophily, direct and undirect). We summarize our contributions as follows: 
\begin{itemize}
\item We define two types of new GNN layers by introducing $p$-Laplacian regularizers to both decomposed and reconstructed framelets. This paves the way for introducing more general Bregman divergence regularization in the graph framelet framework;  %Proposed a flexible, regularized framework. 
\item We propose an iterative algorithm to fit the proposed models and explore an alternative algorithm developed from the F-norm LPP;  %\EScomment{What is this specifically?}

\item We prove that the iteration of the proposed algorithm provides a sequence of spectral graph convolutions and diagonal scaling over framelet-filtered graph signals, this gives an deeper explanation of how $p$-Laplacian regularizer interacts with the framelet.
%a second-time filtering effect on the graph signal that was initially filtered in framelet domain. This illustrates that the incorporation of a ($p$-Laplacian-based) flexible regularizer can restrict the solution of GNN into the region that is more suitable for down-streaming learning tasks. 

\item We connect the proposed $p$-Laplacian regularizer to the previously studied framelet regularizer to illustrate the denoising power of the proposed model. 

\item We investigate the performance of the new models on graph learning tasks for both homophilic (undirected) graphs and heterophilic (directed) graphs. To our best knowledge, we are the first to explore the possibility of applying the framelet GCN for directed graphs. %, under the assumption of Chebyshev polynomial approximation. 
The experiment results demonstrate the effectiveness of pL-UFG and pL-fUFG on real-world node classification tasks with strong robustness.
\end{itemize}

%\textit{Organization.}
The rest of the paper is organized as follows. In Section \ref{Sec:2},
we introduce the $p$-Laplacian operator and regularized GNN, followed by a review of recent studies on regularized graph neural networks, which include implicit layers and graph homophily. In Section \ref{Sec:4}, we introduce the fundamental properties of graph framelet and 
propose the $p$-Laplacian regularized framelet models. Furthermore, we develop an algorithm to solve the regularization problem that is more general than $ p$-Laplacian regularization. We also provide theoretical analysis to show how the $p$-Laplacian based regularizer interacts with the graph framelet. 
% the interaction between $p$-Laplacian regularizer and graph framelet in each iteration.
By the end of Section \ref{Sec:4} a brief discussion on the denoising power of the proposed model is presented. Lastly, we present the experimental results and analysis in Section \ref{Sec:5}. Finally, this paper is concluded in Section \ref{Sec:6}.

\section{Preliminaries} \label{Sec:2}
%\subsection{Graph Neural Network}
\EScomment{This section provides an in-depth exploration of the fundamental concepts, encompassing graphs, graph framelets, and regularized graph neural networks. For the sake of reader comprehension and ease of following the intended ideas, each newly introduced model and definition will be accompanied by a succinct review of its developmental history.}

%\EScomment{We shall be consist with several frequently used notations such as : $p$-Laplacian or $p$-Laplacian; $a_{i,j}$ or $a_{i,j}$. Plus I didn't check the literature review in front of each subsection.} \JScomment{$p$-lap has been modified, will check once complete my assignment~}
\subsection{Basic Notations}
%In this paper, the following standard notations will be utilised.
%\textbf{Frobenius norm}.
Let $\mathcal{G} = (\mathcal{V}, \mathcal{E}, W)$ denote a weighted graph, where $\mathcal{V} = \{v_1, v_2, \cdots, v_N \}$ and $\mathcal{E} \subseteq \mathcal{V} \times \mathcal{V}$ represent the node set and the edge set, respectively. ${\bf X} \in \mathbb{R}^{N \times c}$ is the feature matrix for $\mathcal{G}$ with $\{ \mathbf{x}_1, \mathbf{x}_2, \cdots, \mathbf{x}_N \}$ as its rows, and $W = [w_{i,j}] \in \mathbb{R}^{N \times N}$ is the weight matrix on edges with $w_{i,j}>0$ if $(v_i, v_j) \in \mathcal{E}$ and $w_{i,j} = 0$ otherwise. For undirected graphs, we have $w_{i,j} = w_{j,i}$ which means that $W$ is a symmetric matrix. For directed graphs, it is likely that $w_{i,j} \neq w_{j, i}$ which means that $W$ may not be a symmetric matrix.  In most cases, the weight matrix is the graph adjacency matrix, i.e., $ w_{i,j} \in \{0, 1\}$ with elements $w_{i,j}=1$  if $(v_i, v_j) \in \mathcal{E}$ and $w_{i,j} = 0$ otherwise.  %\AHcomment{(Would it be possible to only use $W$?)}\JScomment{done~} 
Furthermore, let $\mathcal{N}_i = \{v_j: (v_i,v_j)\in \mathcal{E}\}$ denote the set of neighbours of node $v_i$ and $\mathbf{D} = \text{diag}(d_{1,1}, ..., d_{N,N})\in \mathbb{R}^{N\times N}$ denote the diagonal degree matrix with $d_{i,i} = \sum^N_{j=1}w_{i,j}$ for $i = 1, . . . , N$. The normalized graph Laplacian is defined as $\widetilde{\mathbf L} = \mathbf I - \mathbf D^{-\frac12} (\mathbf W + \mathbf I) \mathbf D^{-\frac12}$. Lastly, for any vector $\mathbf x = (x_1, ..., x_c)\in\mathbb{R}^c$, we use $\|\mathbf x\|_2 = (\sum^c_{i=1} x^2_i)^{\frac12}$ to denote its L$_2$-norm, and similarly for any matrix $\mathbf M = [m_{i,j}]$, $\|\mathbf M\|:=\|\mathbf M\|_F = (\sum_{i,j}m^2_{i,j})^{\frac12}$ is used to denote its Frobenius norm.

\subsection{Consistency in Graphs}
%\GaoC{Let us leave Framelet Section 3. We use this subsection for Graph types.} 
Generally speaking, most GNN frameworks are designed under the homophily assumption in which the labels of nodes and neighbours in the graph are mostly identical. The recent work by \cite{ZhuYanZhao2020} emphasises that the general topology GNN fails to obtain outstanding results on the graphs with different class labels and dissimilar features in their connected nodes, which we call heterophilic or low homophilic graphs.
The definition of homophilic and heterophilic graphs are given by:
\begin{definition}[Homophily and Heterophily \citep{FuZhaoBian2022}]
\label{HomophilyHeterophily}
The homophily or heterophily of a network is used to define the relationship between labels of connected nodes. The level
of homophily of a graph can be measured by $\mathcal{H(G)} = \mathbb{E}_{i \in \mathcal{V}}[\vert \{j\}_{j \in \mathcal{N}_{i,y_i=y_i}}\vert/\vert\mathcal{N}_i \vert]$, where 
$\vert\{j\}_{j \in \mathcal{N}_{i,y_i= y_i}}\vert$ denotes the number
of neighbours of $i \in V$ that share the same label as $i$ such that $y_i = y_j$.  $\mathcal{H(G)} \rightarrow 1$ corresponds to strong homophily
while $\mathcal{H(G)} \rightarrow 0$ indicates strong heterophily. We say that
a graph is a homophilic (heterophilic) graph if it has strong
homophily (heterophily).

\end{definition}

% \begin{comment}
% \begin{remark}\label{model_preference}
% The graph homophilic assumption produce a challenge to GNN models as the class of heterophilic graphs contains instances where graph nodes should be sharpened rather than smoothed out. Accordingly, an ideal GNN model must accommodate both homophily and heterophilic graph features.  
% \end{remark}
% \EScomment{I excluded the remark 1 as similar content has been mentioned in introduction. }
% \end{comment}
%\subsection{High frequency dominate}

\subsection{$p$-Laplacian Operator}
The first paper that explores the notation of graph $p$-Laplacian and utilizes it in the regularization problem in graph structure data can be traced back to the work \citep{ZhouSchoelkopf2005}, in which the regularization scheme was further explored to build a flexible GNN learning model in the study \citep{FuZhaoBian2022}. In this paper, we refer to the notation similar to the one used in the paper \citep{FuZhaoBian2022} to define the graph $p$-Laplacian operator. We first define the graph gradient as follows: 
%To understand the $p$-Laplacian operation, it is essential to know the concept of graph gradient and graph divergence.
\begin{definition}[Graph Gradient]
Let $\mathcal{F_V}:=\{\mathbf F \vert \mathbf F: \mathcal V \rightarrow \mathbb R^d\}$ \text{and} $\mathcal{F_E}:=\{\mathbf g \vert\mathbf g: \mathcal{E} \rightarrow \mathbb {R}^d\}$
be the function space on nodes and edges, respectively. %and similarly $\mathcal{F_E}$ the function space on edges. 
Given a graph $\mathcal{G} = (\mathcal{V}, \mathcal{E}, W)$ and a function $\mathbf F\in \mathcal{F_V}$, the graph gradient is an operator $\nabla_W $:$\mathcal{F_V} \rightarrow \mathcal{F_E }$ defined as for all $[i,j] \in \mathcal{E}$,\[(\nabla_W \mathbf F) ([i,j]) : = \sqrt{\frac{w_{i,j}}{d_{j,j}}}\mathbf f_j - \sqrt{\frac{w_{i,j}}{d_{i,i}}}\mathbf f_i.\]   
where $\mathbf f_i$ and $\mathbf f_j$ are the signal vectors on nodes $i$ and $j$, i.e., the rows of $\mathbf F$.
\end{definition}   

%Note that it is straightforward to extend the definition to multiple channel graph signals $\mathbf F$.

Without confusion, we will simply denote $\nabla_W$ as $\nabla$ for convenience. For $[i,j] \notin \mathcal{E}$, $(\nabla \mathbf F)([i,j]) := 0$. The graph gradient of %a function/graph signal 
$\mathbf F$ 
% \footnote{Without loss of generality, here $f$ can be any (vector function) $f: \mathcal V \rightarrow \mathbb R^c$. For example, one can let $f$ be the function that produces the graph signal. } 
at a vertex $i$, $\forall i \in \{1, ..., N\}$, is defined as $\nabla \mathbf F(i) :=
((\nabla \mathbf F)([i, 1]);\dots ; (\nabla \mathbf F)([i,N]))$ and its Frobenius norm is given by $\|\nabla \mathbf F(i) \|_2 := (\sum ^N_{j=1}(\nabla \mathbf F)^2([i,j]))^{\frac{1}{2}}$ which measures the variation of $\mathbf F$ around node $i$. Note that we have two explanations for the notation $\nabla \mathbf F$: one as a graph gradient (over edges) and one as node gradient (over nodes). The meaning can be inferred from its context in the rest of the paper.  %Analogized by the Laplacian operator in the continuous setting, which is the divergence of the gradient of a continuous function, here we provide the definition of graph divergence: 
We also provide the definition of graph divergence, analogous to the Laplacian operator in continuous setting, which is the divergence of the gradient of a continuous function.
\begin{definition}[Graph Divergence]
Given a graph $\mathcal{G} = (\mathcal{V}, \mathcal{E}, W)$ and a function $\mathbf F: \mathcal V \rightarrow \mathbb R^d$, $\mathbf g: \mathcal E \rightarrow \mathbb R^d$, the graph divergence is an operator $\text{div}: \mathcal F_\mathcal E \rightarrow \mathcal F_\mathcal V$ which satisfies: 
\begin{align}
    \langle \nabla \mathbf F,\mathbf g \rangle = \langle \mathbf F, -\text{div}(\mathbf g) \rangle.
\end{align}
Furthermore, the graph divergence can be computed by: 
\begin{align}
    \text{div}(\mathbf g)(i) = \sum_{j =1 }^N \sqrt{\frac{w_{i,j}}{d_{i,i}}}(\mathbf g[i,j]-\mathbf g[j,i]).
\end{align}
\end{definition}

% We measure
% the variation of $f$ over the whole graph $\mathcal{G}$ by $\mathcal{S}_p(f)$ where
% it is defined as for $p \geq 1$,
% \begin{align}
%     \mathcal{S}_p(f) &:= \frac{1}{2} \sum_{i=1}^N \sum_{j=1}^N \|(\nabla f)([i,j])\|^p\\
%     & = \frac{1}{2} \sum_{i=1}^N \sum_{j=1}^N \left\| \sqrt{\frac{W_{i,j}}{D_{j,j}}}f(j) - \sqrt{\frac{W_{i,j}}{D_{i,i}}}f(i) \right\|^p.\\
% \end{align}

Given the above definitions on graph gradient and divergence, we reach the definition of the graph $p$-Laplacian.
\begin{definition}[$p$-Laplacian operator \citep{FuZhaoBian2022}]
    Given a graph $\mathcal{G} = (\mathcal{V}, \mathcal{E}, W)$ and a multiple channel signal function $\mathbf F: \mathcal V \rightarrow \mathbb R^d$, the graph $p$-Laplacian is an operator $\Delta_p: \mathcal F_\mathcal V \rightarrow \mathcal F_\mathcal V$, defined by:
\begin{align}\label{p_lap}
    \Delta_p \mathbf F: = - \frac{1}{2} \text{div}( \|\nabla \mathbf F\|^{p-2} \nabla \mathbf F), \quad \text{for} \,\,\, p\geq 1.
\end{align}
where $\|\cdot\|^{p-2}$ is element-wise power over the node gradient $\nabla \mathbf F$.
\end{definition}

\begin{remark}\label{p_1_explain}
Clearly, when we have $p=2$, Eq. \eqref{p_lap} recovers the classic graph Laplacian. When $p =1$, we can analogize Eq. \eqref{p_lap} as a curvature operator defined on the nodes of the graph, because when $p =1$, we have Eq. \eqref{p_lap} as $\Delta_1 \mathbf F = -\frac12\text{div}\left(\frac{\nabla \mathbf F}{\|\nabla \mathbf F\|}\right)$. 
% \begin{align}
%     \Delta_1 f = \text{div}\left(\frac{\nabla f}{\|\nabla f\|}\right)
% \end{align}
This aligns with the definition of mean curvature operator defined on the continuous domain. Furthermore, we note that the $p$-Laplacian operator is linear when $p = 2$, while in general for $p \neq 2 $, the $p$-Laplacian is a non-linear operator since $\Delta_p(a\mathbf F) \neq a\Delta_p(\mathbf F) \,\,\, \text{for} \,\,\, a \in \mathbb R\backslash \{1\}$.
\end{remark}

\begin{definition}[$p$-eigenvector and $p$-eigenvalue]
Let $\psi_p(\mathbf v) = (\| v_1 \|^{p-2} v_1, ..., \allowbreak \| v_N \|^{p-2} v_N)$ for $\mathbf v \in \mathbb R^N$. With some abuse of the notation, we call $\mathbf u\in \mathcal F_V$ ($d=1$) a $p$-eigenvector of $\Delta_p$ if $\Delta_p \mathbf u = \lambda \psi_p(\mathbf u)$ where $\lambda$ is the associated $p$-eigenvalue.
\end{definition}

Note that in above definition, we use the fact that $\mathbf u$ acting on all nodes gives a vector in $\mathbb R^N$. Let $\psi_p(\mathbf U) = (\psi_p(\mathbf U_{;,1}), ..., \psi_p(\mathbf U_{:, N}))$ for $\mathbf U \in \mathbb R^{N \times N}$. One \citep{FuZhaoBian2022} can show that $p$-Laplacian can be decomposed as $\Delta_p = \psi_p(\mathbf U) \mathbf \Lambda \psi_p(\mathbf U)^T$ for some diagonal matrix $\boldsymbol{\Lambda}$.

\subsection{Graph Framelets}
Framelet is a type of wavelet frame.  The study by \cite{Sweldens1998} is the first to present a wavelet with a lifting scheme that provides a foundation in the research of wavelet transform on graphs. With the increase of computational power, \cite{HammondVandergheynstGribonval2009} proposed a framework for the wavelet transformation on graphs and employed Chebyshev polynomials to make approximations on wavelets. \cite{Dong2017} further developed %high computational 
tight framelets on graphs. % by approximating smooth function in tiny slice with proper filtered Chebyshev polynomial approximation. 
The new design has been applied for graph learning tasks % extended into graphs
\citep{ZhengZhouGaoWangLioLiMontufar2021} with great performance enhancement against the classic GNNs. The recent studies show %presenting the Framelet with graph neural network have achieved outstanding empirical study results, such that \citep{ZhengZhouGaoWangLioLiMontufar2021} who pointing out 
that framelet can naturally decompose the graph signal and re-aggregate them  effectively, achieving a significant result on graph noise reduction \citep{ZhouLiZhengWangGao2021}  with the use of a double term regularizer on the framelet coefficients. % to present a robust optimization technique for graph denoising. 
Combing with singular value decomposition (SVD), the framelets have been made applicable to directed graphs \citep{ZouHanLinGao2022}. 

A simple method of building more versatile and stable framelet families was suggested by \cite{yang2022quasi} which is known as Quasi-Framelets. In this study, we will introduce graph framelets using the same architecture described in the paper \citep{yang2022quasi}. We begin by defining the filtering functions for Quasi-Framelets: 

\begin{definition}[Filtering functions for Quasi-Framelets] We call a set of $R+1$ positive filtering functions defined on $[0, \pi]$, $\mathcal{F} = \{g_0(\xi), g_1(\xi), ..., g_R(\xi)\}$, Quasi-Framelet scaling functions if the following identity condition is satisfied:
\begin{align}
g_0(\xi)^2 + g_1(\xi)^2 + \cdots + g_R(\xi)^2 \equiv 1,\;\;\; \forall \xi \in [0, \pi], \label{eq:4}
\end{align}
such that $g_0$ descends from 1 to 0 and $g_R$ ascends from 0 to 1 as frequency increases over the spectral domain $[0, \pi]$. 
\end{definition}
Particularly $g_0$ aims to regulate the highest frequency while $g_R$ to regulate the lowest frequency, and the rest to regulate other frequencies in between.

Consider a graph $\mathcal{G} =(\mathcal{V}, \mathcal{E})$ with its normalized graph Laplacian $\widetilde{\mathbf L}$. Let $\widetilde{\mathbf L}$ have the eigen-decomposition $\widetilde{\mathbf L} = \mathbf U\Lambda \mathbf U^T$ where $\mathbf U$ is the orthogonal spectral bases with its spectra $\Lambda = \text{diag}(\lambda_1, \lambda_2, ..., \lambda_N)$ in increasing order. For a given set of Quasi-Framelet functions $\mathcal{F} = \{g_0(\xi), g_1(\xi), ..., g_R(\xi)\}$ defined on $[0, \pi]$ and a given level $J$ ($\geq 0$), define the following Quasi-Framelet signal transformation matrices

\begin{align}
    \mathcal{W}_{0,J} &= \mathbf U g_0(\frac{\boldsymbol{\Lambda}}{2^{m+J}}) \cdots g_0(\frac{\boldsymbol{\Lambda}}{2^{m}}) \mathbf U^T, \label{eq:8a}\\
    \mathcal{W}_{r,0} &= \mathbf U g_r(\frac{\boldsymbol{\Lambda}}{2^{m}}) \mathbf U^T, \;\;\text{for } r = 1,...,R, \label{eq:8b}\\
    \mathcal{W}_{r,\ell} &= \mathbf U g_r(\frac{\boldsymbol{\Lambda}}{2^{m+\ell}})g_0(\frac{\boldsymbol{\Lambda}}{2^{m+\ell-1}}) \cdots g_0(\frac{\boldsymbol{\Lambda}}{2^{m}}) \mathbf U^T,\label{eq:8} \\ 
    \;\;\;\; & \text{for }  r = 1,...,R, \ell = 1,...,J.\notag
\end{align}

Note that in the above definition, $m$ is called the coarsest scale level which is the smallest $m$ satisfying $2^{-m}\lambda_N \leq \pi$. Denote by   $\mathcal{W} = [\mathcal{W}_{0,J}; \mathcal{W}_{1,0}; ...; \mathcal{W}_{R,0}; $ $\mathcal{W}_{1,1}; ..., \mathcal{W}_{R,J}]$ as the stacked matrix.  It can be proved that $\mathcal{W}^T\mathcal{W} = \mathbf I$, thus providing a signal decomposition and reconstruction process based on $\mathcal{W}$. We call this graph Framelet transformation.

In order to alleviate the computational cost imposed by eigendecomposition for the graph Laplacians, the framelet transformation matrices can be approximated by Chebyshev polynomials. Empirically, the implementation of the Chebyshev polynomials $\mathcal{T}^n_r(\xi)$ with a fixed degree $n$, $n=3$ is sufficient to approximate $g_r(\xi)$.
%Suppose we approximate $g_j(\xi)$ by Chebyshev polynomials $\mathcal{T}^n_j(\xi)$ of a fixed degree $n$. In practice, $n=3$ is good enough. 
In the sequel, we will simplify the notation by using  $\mathcal{T}_r(\xi)$ instead of $\mathcal{T}^n_r(\xi)$. Then the Quasi-Framelet transformation matrices are defined in Eqs. \eqref{eq:8b} - \eqref{eq:8} can be approximated by, %{for } $k=1, ..., R, \ell = 1, ..., L$,
\begin{align}
    \mathcal{W}_{0,J} &\approx %\mathbf U \mathcal{T}_0(\frac{\boldsymbol{\Lambda}}{2^{L+m}}) \cdots \mathcal{T}_0(\frac{\boldsymbol{\Lambda}}{2^{m}}) \mathbf U^T =
    \mathcal{T}_0(\frac1{2^{m+J}}\widetilde{\mathbf L}) \cdots \mathcal{T}_0(\frac{1}{2^{m}}\widetilde{\mathbf L}),   \label{eq:Ta0}\\
    %\;\;\;\;\;\;
    \mathcal{W}_{r,0} &\approx %\mathbf U \mathcal{T}_k(\frac{\boldsymbol{\Lambda}}{2^{m}}) \mathbf U^T =  
    \mathcal{T}_r(\frac1{2^{m}}\widetilde{\mathbf L}),  \;\;\; \text{for } r = 1, ..., R, \label{eq:Ta}
 \\ 
    \mathcal{W}_{r,\ell} &\approx %\mathbf U \mathcal{T}_k(\frac{\boldsymbol{\Lambda}}{2^{m+\ell}})\mathcal{T}_0(\frac{\boldsymbol{\Lambda}}{2^{m+\ell-1}}) \cdots \mathcal{T}_0(\frac{\boldsymbol{\Lambda}}{2^{m}}) \mathbf U^T  = 
    \mathcal{T}_r(\frac{1}{2^{m+\ell}}\widetilde{\mathbf L})\mathcal{T}_0(\frac{1}{2^{m+\ell-1}}\widetilde{\mathbf L}) \cdots \mathcal{T}_0(\frac{1}{2^{m}}\widetilde{\mathbf L}), \label{eq:Tc}\\
    &\;\;\;\; \text{for } r=1, ..., R, \ell= 1, ..., J. \notag
\end{align}

\textit{Remark 1:} The approximated transformation matrices defined in Eqs. \eqref{eq:Ta0}-\eqref{eq:Tc} simply depend on the graph Laplacian. For directed graphs, we directly take in the Laplacian normalized by the out degrees in our experiments. We observe this strategy leads to improved performance in general. 

For a graph signal $\mathbf{X}$, the
framelet (graph) convolution similar to the spectral graph convolution that can be defined as
\begin{align}
\theta \star \mathbf{X} = \mathcal{W}^T(\textrm{diag}(\theta))(\mathcal{W}\mathbf X), \label{Eq:9new}
\end{align}
where $\theta$ is the learnable network filter.  We also call $\mathcal{W}\mathbf X$ the framelet coefficients of $\mathbf X$ in  Fourier domain. The signal then will be filtered in its spectral-domain according to learnable filter $\textrm{diag}(\theta)$.

\subsection{Regularized Graph Neural Network}
In reality, graph data normally have  large, noisy and  complex structures \citep{LeskovecFaloutsos2006,HeKempe2015}. This brings the challenge of choosing a suitable neural network for fitting the data. It is well known that one of the common computational issues for most classic GNNs is over-smoothing which 
can be quantified by Dirichlet energy that converges to zero as the number of layers increases. This observation leads to an investigation into the so-called implicit layers on regularized graph neural networks. %\GaoC{Dont understand the meaning of this!}\WangC{rewrote:)} . 

The first paper to consider GNNs layer as a regularized signal smoothing process is done by \cite{ZhuWangShiJiCui2021}, in which the classic GNN layers are interpreted as the solution of the regularized optimization problem, with certain approximation strategies to avoid matrix inversion in the closed-form solution. The regularized layer also can be linked to the implicit layer \citep{ZhangGuMateusz2020}, more specific examples are given by \cite{ZhuWangShiJiCui2021}. In general, given the node features, the output of the GNN layer can be written as the solution for the following optimization problem
\begin{align}
\mathbf F = \argmin_{\mathbf F} \left\{\mu\|\mathbf X - \mathbf F\|^2_F +\frac{1}2 \text{tr}(\mathbf F^T \widetilde{\mathbf L}\mathbf F)\right\}, \label{Eq:1new}
\end{align}
where $\mathbf F$ is the graph signal. Eq. \eqref{Eq:1new} defines the layer output as the solution for the (layer) optimization problem with a regularization term $\text{tr}(\mathbf F^T \widetilde{\mathbf L}\mathbf F)\}$, which is able to enforce smoothness of a graph signal. The closed form
solution is given by $\mathbf F = (\mathbf I + \frac1{2\mu} \widetilde{\mathbf L})^{-1}\mathbf X$. %\AHcomment{(can we remove the approximation part?)}  \approx  (\mathbf I - \frac1{2\mu} \widetilde{\mathbf L})\mathbf X. 
It is computationally inefficient because of matrix inversion. However, we can rearrange it to $(\mathbf I + \frac1{2\mu} \widetilde{\mathbf L})\mathbf F = \mathbf X$ and interpret it as a fixed point solution to $\mathbf F^{(t+1)} = -\frac1{2\mu} \widetilde{\mathbf L} \mathbf F^{(t)} + \mathbf X$. The latter is then the implicit layer in GNN at layer $t$. 
%Since it can be interpreted as an implicit layer defined by $(\mathbf I + \frac1{2\mu} \widetilde{\mathbf L})\mathbf F = \mathbf X$, which enable the equation converted into a fixed-point iterate as $\mathbf F^{(t+1)} = -\frac1{2\mu} \widetilde{\mathbf L} \mathbf F^{(t)} + \mathbf X$ in each layer. 
Such an iteration is motivated by the recurrent GNNs as shown in several recent works \citep{GuChangZhuSojoudiElGhaoui2020,LiuKawaguchiHooiWangXiao2021,ParkChooPark2021}.  Different from explicit GNNs, the output features $\mathbf F$ of a general implicit GNN are directly modeled as the solution of a well defined implicit function, e.g., the first order condition from an optimization problem. Denote the rows of $\mathbf F$ by $\mathbf f_i$ ($i=1, 2, ..., N$) in column shape.  It is well known that the regularization term \eqref{Eq:1new} is
\[
\frac12\text{tr}(\mathbf F^T \widetilde{\mathbf L}\mathbf F) = \frac12 \sum_{(v_i,v_j)\in\mathcal{E}}\left\|\sqrt{\frac{w_{i,j}}{d_{j,j}}}\mathbf f_j -\sqrt{\frac{w_{i,j}}{d_{i,i}}}\mathbf f_i\right\|^2,
\]
which is the so-called graph Dirichlet energy \citep{ZhouSchoelkopf2005}. As proposed in the work \citep{ZhouSchoelkopf2005}, one can replace the graph Laplacian matrix in the above equation to the pre-defined graph $p$-Laplacian, then the $p$-Laplacian based energy denoted as $\mathcal S_p(\mathbf F)$ can be defined as (for any $p \geq 1$):

% Then in the recent paper \citep{FuZhaoBian2022} the authors proposed using the so-called $p$-Laplacian regularizer, in particular, for any given (weighted) finite graph $\mathcal{G} = (\mathcal{V}, \mathcal{E}, \mathbf{W})$ the graph and the functions 

% and a function $f \rightarrow: \mathcal V \rightarrow \mathbb R$( in this paper, one can consider $f$ as  the family of function that produce graph signal), the graph $p$-Laplacian is an operator $\Delta_p: \mathcal F_\mathcal V \rightarrow \mathcal F_\mathcal V$ which is defined by: 
% \begin{align}
%     \Delta_p f : = \frac{1}{2} \text{div}\left(\|\nabla f\|^{$p$-2}\nabla f\right) \quad \text{for} $p$ \geq 1
% \end{align}
% where $\text{div}$ is the graph divergence operator with the form: $\text{div} : = \sum_{j=1}^N \sqrt{\frac{W_{i,j}}{D_{i,i}}(\mathbf f_i - \mathbf f_j)}$.

% as defined below, for any $p\geq 1$,
\begin{align}
\mathcal{S}_p(\mathbf F) = \frac12\sum_{(v_i,v_j)\in\mathcal{E}}\left\|\sqrt{\frac{w_{i,j}}{d_{j,j}}}\mathbf f_j -\sqrt{\frac{w_{i,j}}{d_{i,i}}}\mathbf f_i\right\|^p, \label{Eq:Lp}
\end{align}
where we adopt the definition of element-wise $p$-norm as in paper \citep{FuZhaoBian2022}. Finally, the regularization problem becomes
\begin{align}
\mathbf F = \argmin_{\mathbf F} \left\{\mu\|\mathbf X - \mathbf F\|^2_F + \mathcal{S}_p(\mathbf F)\right\}. \label{Eq:2}
\end{align} 
\EScomment{With strong generalizability of the regularization form in Eq.~\eqref{Eq:2}, it is natural to consider to deploy Eq.~\ref{Eq:2} to multi-scale graph neural networks (i.e., graph framelets) to explore whether the benefits of allocating $p$-Laplacian based regularizer still remains and how the changes of $p$ within included regularizer interacts with the feature propagation process via each individual filtered domains. In the next section, we will accordingly proposed our model and provide detailed discussion and analysis on it. }

\section{The Proposed Models}\label{Sec:4}
\begin{comment}
    \begin{figure*}[t]
    \centering
    \includegraphics[scale=0.3]{framelet.png}
    \caption{The above figure shows the framelet framework by giving a graph with structure (adjacency matrix) and feature information.} 
    \label{fig:denoise_cora}
\end{figure*}
\end{comment}

%\JScomment{Will add graph gradient in the framework}

\EScomment{In this section, we show our proposed models: $p$-Laplacian Framelet GNN (pL-UFG) and $p$-Laplacian Fourier Undecimated Framelet GNN (pL-fUFG) models. In addition, we introduce a more general regularization framework and describe the algorithms of our proposed models.}

\subsection{$p$-Laplacian Undecimated Framelet GNN (pL-UFG)} 
% The layer of framelet GNN is derived from Eq. \eqref{Eq:9new}. 

\EScomment{Instead of simply taking the convolution result as the framelet layer output (derived in Eq.~\eqref{Eq:9new}), we apply the $p$-Laplacian regularization on the framelet reconstructed signal by imposing the following optimization to define the new layer, } %First we apply the idea of L$p$-GCN over the framelet filtered graph signal as
%\GaoC{We will decide whether use $\mu$ or $\lambda$!}
\begin{align}
\mathbf F = \argmin_{\mathbf{F}}\mathcal{S}_p(\mathbf{F}) + \mu \|\mathbf{F} - \mathcal{W}^T\textrm{diag}(\theta)\mathcal{W}\mathbf X\|^2_F \label{Eq:1},
\end{align} 
where $\mu \in (0, \infty)$. %, the input graph feature $\mathbf{X} =  (\mathbf{X}^T_{1,:}, ..., \mathbf{X}^T_{N,:})^T \in \mathbb{R}^{N \times c}$ of graph $G$ with $N$ nodes, for $i \in [N]$, $\mathbf{X}_{i,:} \in \mathbb{R}^{1 \times c}$ and $\mathbf{F} = (\mathbf{F}^T_{1,:}, ..., \mathbf{F}^T_{N,:})^T \in \mathbb{R}^{N \times c}$ be the function value of graph signal $\mathbf{f}$, the $i^{th}$ row vector $\mathbf{F}_{i,:}\in\mathbb{R}^{1 \times c}$, $i\in[N]$. 
In which we recall that the first term $\mathcal{S}_p(\mathbf{F})$ in the above equation is the $p$-Laplacian based energy defined in Eq. \eqref{Eq:Lp}. $\mathcal{S}_p(\mathbf{F})$ measures the total signals' variation throughout the graph-based format of $p$-Laplacian.  
The second term dictates that optimal $\mathbf{F}$ should not deviate excessively from the framelet reconstructed signal for the input signal $\mathbf{X}$. % with framelet transformation $\mathcal{W}$, 
%The learnable filter diag($\theta$)vector of \textit{network filter} $\theta$\GaoC{need double check!}. 
Each component of the network filter $\theta$ in the frequency domain is applied to the framelet coefficients  $\mathcal{W}\mathbf{X}$. We call the model in Eq.\eqref{Eq:1} pL-UFG2. 

One possible variant to model \eqref{Eq:1} is to apply the regularization individually on the reconstruction at each scale level, i.e., for all the $r, \ell$, define
\begin{align}
% \begin{aligned}
 \mathbf{F}_{r, \ell}  &= 
\argmin_{ \mathbf{F}_{r, \ell} } 
\mathcal{S}_p( \mathbf{F}_{r, \ell} ) + \mu \| \mathbf{F}_{r, \ell} - \mathcal{W}^T_{r, \ell}  \textrm{diag}(\theta_{r, \ell}) \mathcal{W}_{r, \ell}  \mathbf X \|^2_F, \nonumber\\
\mathbf F &= \mathbf F_{0,J} + \sum^R_{r=1}\sum^J_{\ell=0} \mathbf F_{r, \ell}.
% \end{aligned}
\label{Eq:1variant}
\end{align} 
We name the model with the propagation in the above equation as pL-UFG1. In our experiment, we note that  pL-UFG1  performs better than pL-UFG2. 

%%%%%  We dont need this details
\begin{comment}
\begin{definition}[Graph Gradient]
    Given a graph $\mathcal{G} = (\mathcal{V}, \mathcal{E})$ and a function $f: \mathcal{V} \rightarrow \mathbb{R}$, the graph gradient is an operator $\nabla : \mathcal{F}_\mathcal{V} \rightarrow \mathcal{F}_\mathcal{E}$ defined as for all $[i.j] \in \mathcal{E}$,
    \begin{align} 
    (\nabla f)([i,j]) : = \sqrt{\frac{W_{i,j}}{D_{j,j}}} f(j) - \sqrt{\frac{W_{i,j}}{D_{i,i}}} f(i). \label{Eq:2}
\end{align}

For $[i,j] \notin \mathcal{E}, (\nabla f)([i,j]) : = 0$. The graph gradient of function $f$ at a vertex $i,i \in [N]$ is defined as $\nabla f(i):= (\nabla f)([i,1]), ..., (\nabla f)([i,N])$ and its Frobenius norm is given by $||\nabla f(i)||_2 := (\sum^N_{j=1} (\nabla f)^2 ([i,j]))^{\frac{1}{2}}$, which measures the variation of $f$ around node $i$. The measurement of variation on $f$ over the whole graph $\mathcal{G}$ by $\mathcal{S}_p$ that is defined as for $p \geq 1$,

\begin{align}
    \mathcal{S}_p(f) : &= \frac{1}{2} \sum_{i=1}^N \sum_{i=1}^N \| (\nabla f)([i,j]) \|^p\\
    & = \frac{1}{2} \sum_{i=1}^N \sum_{i=1}^N \left|\left| \sqrt{\frac{W_{i,j}}{D_{j,j}}} f(j) - \sqrt{\frac{W_{i,j}}{D_{i,i}}} f(i) \right|\right|^p. \label{Eq:3}
\end{align}
\end{definition}
\end{comment} 
%%% end of comments

\subsection{$p$-Laplacian Fourier Undecimated Framelet GNN (pL-fUFG)}
In pL-fUFG, we take a strategy of regularizing the framelet coefficients. %reconstructed signal. 
Informed by earlier experience, we  consider the following optimization problem for each framelet transformation in Fourier domain. For each framelet transformation $\mathcal{W}_{r,j}$ defined in Eqs. \eqref{eq:Ta}-\eqref{eq:Tc}, define 
\begin{align}
 \mathbf{F}_{r,\ell}  = 
\argmin_{ \mathbf{F}_{r,\ell} } 
\mathcal{S}_p( \mathbf{F}_{r,\ell} ) + \mu \| \mathbf{F}_{r,\ell} - \textrm{diag}(\theta_{r,\ell}) \mathcal{W}_{r,\ell}  \mathbf X \|^2_F. 
\label{Eq:4}
\end{align} 
Then the final layer output is defined by the reconstruction as follows
\begin{align}
    \mathbf F = \mathcal{W}^T_{0, J} \mathbf{F}_{0,J} + \sum^R_{r=1}\sum^J_{\ell =0}  \mathcal{W}^T_{r, \ell} \mathbf{F}_{r, \ell}. 
\label{Eq:5}
\end{align}
Or we can only aggregate all the regularized and filtered framelet coefficients in the following way
\begin{align}
    \mathbf F =  \mathbf{F}_{0,J} + \sum^R_{r=1}\sum^J_{\ell =0}    \mathbf{F}_{r,\ell}. 
\end{align}
\EScomment{With the form of both pL-UFG and pL-fUFG, in the next section we show an even more generalized regularization framework by incorporating $p$-Laplacian with graph framelets.}

%\GaoC{We shall draw an architecture diagram here.}

%where the first term of equation \ref{Eq:4} represent the optimal signal.$\mathbf{F}_{i,:} \in \mathbb{R}^{1\times c}, i \in \mathbf{N}$ the function value of $f$ at $i$th vertex in graph $\mathcal{G}$, $\mathbf{F} = (F_{1,:}^T \dots F_{N,:}^T)$,        
\subsection{More General Regularization}\label{More General Regularization}
\begin{figure*}[t]
    \centering
    \includegraphics[scale=0.3]{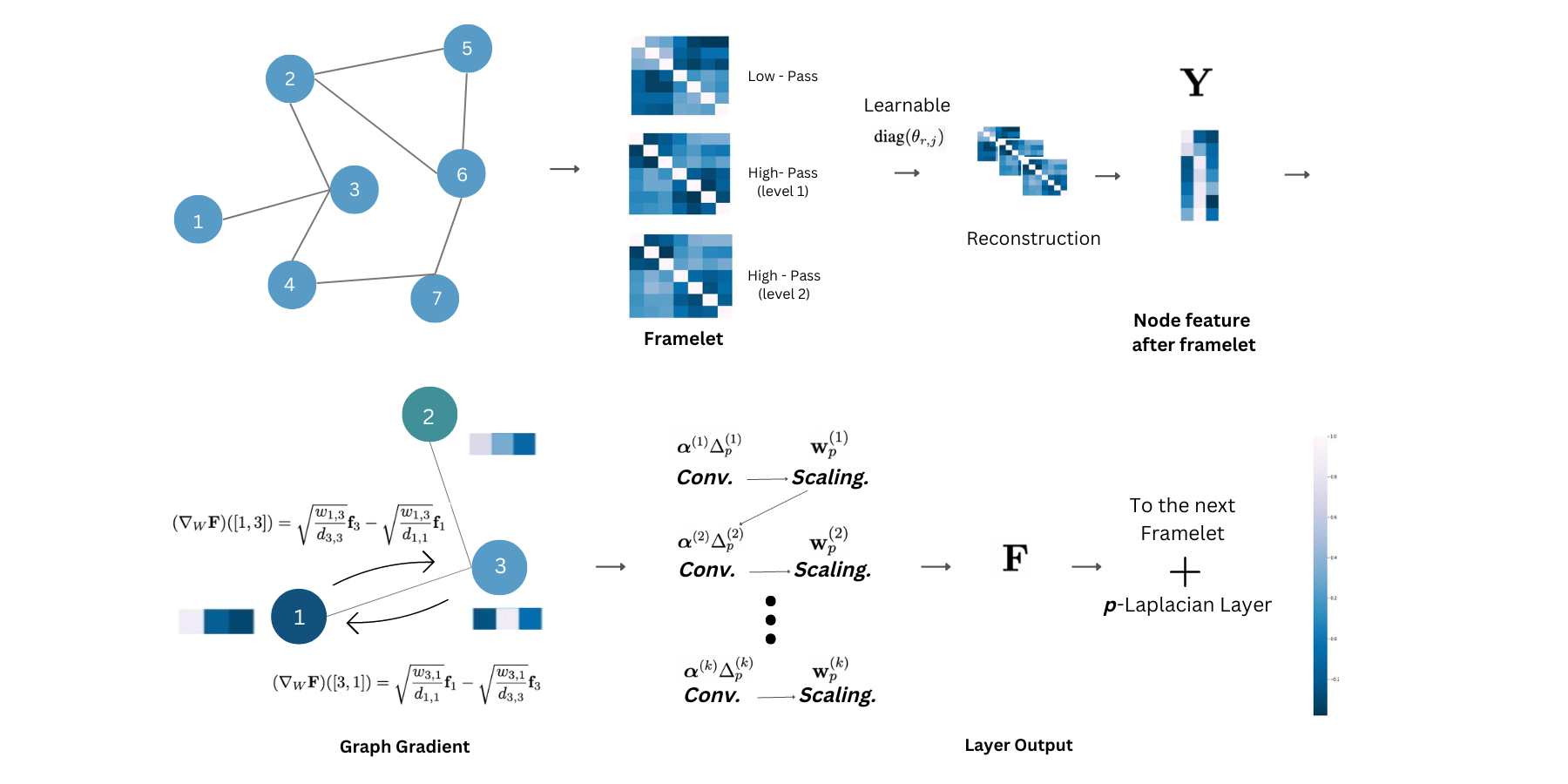}
    \caption{
    The figure above shows the working flow of the $p$-Laplacian regularized framelet. The input graph data is first filtered and reconstructed by the framelet model which contains one low-pass and two high-pass filters (i.e., $J=2$). Then, the result (denoted as $\mathbf Y$) is further regularized by a sequence of graph convolution and diagonal rescaling induced by the $p$-Laplacian, which is generated based on the graph gradient information, serving as an implicit layer of the model. By adjusting the $p$ value, node features resulting from this implicit layer can be smoothed or sharpened accordingly, thus making the model adopt both homophily and heterophilic graphs. Lastly, the layer output will then be either forwarded to the task objective function or to the next framelet and $p$-Laplacian layers before the final prediction task. 
    % The above figure shows that the $p$-Laplacian-based regularization framework interacts with the framelet. The example graph is a heterophilic graph. The framelet firstly filters node features and then calculates the graph gradient of nodes. The example nodes of 1 and 3 are significantly different which is using two different colors and then passing through  $p$-Laplacian convolution with scaling. The layer output $\mathbf{F}$ is regarded as the new graph feature for the same pipeline in the figure, it can go to the next Framelet and $p$-Laplacian layer or to the final loss function.
    %how the penalty term $\mathcal S_p^\phi$ is built based on graph gradient information, and the right side of the figure shows the differences in terms of the range of solution space (presented as the level set of $\mathbf F$) due to different values of $p$. One can clearly see that a higher penalty (i.e., $p=2$, presented as the bold circle in the middle) term intersects the framelet solution space at the inner circle of its level set. In contrast, a lower penalty term (i.e., $p=1$, presented as a shaded square at the middle ) maintains a higher variation of the framelet solution space by only touching the out circles of the framelet solution level sets.
    } 

    % of $p=1$ is obtained from the outer-circle level only, the model is regularized with a lower penalty term ($p=1$) than $p=2$.
    \label{fig:regu}
\end{figure*}
For convenience, we write the graph gradient for multiple channel signals $\mathbf F$ as 
\[
\nabla_W \mathbf F ([i,j]) =\sqrt{\frac{w_{i,j}}{d_{j,j}}}\mathbf f_j -\sqrt{\frac{w_{i,j}}{d_{i,i}}}\mathbf f_i.
\]
or simply $\nabla \mathbf F$ if no confusion.
% \AHcomment{(the notation is not consistent, we shall use $\nabla_{i,j}$ as before)}
% \EScomment{This may be a typo since the conference version. I also wondering the notation here}
For an undirected graph we have $\nabla_W \mathbf F ([i,j])  = - \nabla_W \mathbf F ([j,i]) $. 
With this notation, we can re-write the $p$-Laplacian regularizer in \eqref{Eq:Lp} (the element-wise $p$-norm) in the following. % \textcolor{red}{[Guo] is $\nabla_{i,j}(w, \mathbf f)$ a scalar? If so, the following derivation is right. Otherwise, the last equality in (20) need to be checked. }\GaoC{In [37] here is elementwise norm.}
\begin{align}
&\mathcal{S}_p(\mathbf F)  = \frac12\sum_{(v_i,v_j)\in\mathcal{E}}\left\|\nabla_W \mathbf F ([i,j])\right\|^p \\ 
%&= \frac12\sum_{v_i\in\mathcal{V}} \sum_{v_j \sim v_i} \left\|\sqrt{\frac{w_{i,j}}{d_{j,j}}}\mathbf f_j -\sqrt{\frac{w_{i,j}}{d_{i,i}}}\mathbf f_i\right\|^p\\
=&\frac12\sum_{v_i\in\mathcal{V}} \left[ \left( \sum_{v_j \sim v_i} \left\|\nabla_W \mathbf F ([i,j])\right\|^p\right)^{\frac1p} \right]^p 
= \frac{1}{2} \sum_{v_i \in \mathcal{V}} \| \nabla_W \mathbf F (v_i)  \|_p^p \notag
\end{align}
where $v_j \sim v_i$ stands for the node $v_j$ that is connected to node $v_i$ and 
$\nabla_W\mathbf F (v_i)= \left( \nabla_W \mathbf F ([i,j]) \right)_{v_j: (v_i, v_j) \in \mathcal{E}}$ is the node gradient vector for each node $v_i$ and 
% \begin{align}
% \mathcal{S}_p(\mathbf F) = \frac12 \sum_{v_i\in\mathcal{V}} \|\nabla_W\mathbf F (v_i)\|^p_p =  \frac12 \sum_{v_i\in\mathcal{V}} \phi(\|\nabla_W\mathbf F (v_i)\|_p)
% \end{align}
$\|\cdot\|_p$ is the vector $p$-norm.  In fact, $\|\nabla_W\mathbf F (v_i)\|_p$ measures the variation of $\mathbf F$ in the neighbourhood of each node $v_i$. Next, we generalize the regularizer by considering any positive convex function $\phi$ as
% Naturally we can use any other positive convex function $\phi$ to penalize large variations of $\mathbf F$. Hence for a given such function $\phi$, we define the following generalized regularizer
\begin{align}
\mathcal{S}^{\phi}_p(\mathbf F) = \frac12 \sum_{v_i\in\mathcal{V}} \phi(\|\nabla_W\mathbf F (v_i)\|_p). \label{function_phi}
\end{align}
It is clear when $\phi(\xi) = \xi^p$, we recover the $p$-Laplacian regularizer. 
In image processing field, several penalty functions have been proposed in the literature. For example, $\phi(\xi) = \xi^2$ is known in the context of Tikhonov regularization \EScomment{\citep{li2022tikhonov,belkin2004tikhonov, assis2021neural}}. When $\phi(\xi) = \xi$ (i.e. $p=1$), it is the classic total variation regularization.  When $\phi(\xi) = \sqrt{\xi^2 + \epsilon^2} - \epsilon$, it is referred as the regularized total variation. \EScomment{An example work can be found in the work \citep{chengwang2020}}. The case of $\phi(\xi) = r^2\log(1 + \xi^2/r^2)$ corresponds to the non linear diffusion \EScomment{\citep{TomoyukiTakayuki2023}}.   

In pL-UFG and pL-fUFG, we use $\mathbf Y$ to denote $\mathcal{W}^T\textrm{diag}(\theta)\mathcal{W}\mathbf X$ and $\,\,\,$
$\textrm{diag}(\theta_{r,j}) \mathcal{W}_{r,j}  \mathbf X$ respectively, as then we propose the generalized $p$-Laplacian regularization models as
\begin{align}
\mathbf F = \argmin_{\mathbf{F}}\mathcal{S}^{\phi}_p(\mathbf{F}) + \mu \|\mathbf{F} - \mathbf Y\|^2_F. \label{Eq:Model3}
\end{align} 

\subsection{The Algorithm}
%\GaoC{I suggest we derive the algorithm from \eqref{Eq:Model3}} 
%\GaoC{We may writ this into a Theorem.}
We derive an iterative algorithm for solving the generalized  $p$-Laplacian regularization problem \eqref{Eq:Model3}, presented as the following theorem.  

% \EScomment{I think here $M_i,j$ should also in the form of $\|\nabla_{i,j}(w, \mathbf f)\|^{$p$-2}$ rather than using the Laplacian operator}

\begin{theorem}\label{algorithm_thm}
For a given positive convex function $\phi(\xi)$, define
\begin{align*}
M_{i,j} = & \frac{w_{i,j}}2  \left\| \nabla_W \mathbf F ([i,j]) \right\|^{p-2} \cdot \left[\frac{\phi'(\|\nabla_W\mathbf F (v_i)\|_p)}{\|\nabla_W\mathbf F (v_i)\|^{p-1}_p} \right. 
 + \left.\frac{\phi'(\|\nabla_W\mathbf F (v_j)\|_p)}{\|\nabla_W\mathbf F (v_j)\|^{p-1}_p} \right],\\
\alpha_{ii}  = & 1/\left(\sum_{v_j\sim v_i}\frac{M_{i,j}}{d_{i,i}} + 2\mu\right),\quad\quad
\beta_{ii} =  2\mu \alpha_{ii}
\end{align*}
and denote the matrices $\mathbf M = [M_{i,j}]$, $\boldsymbol{\alpha} = \text{diag}(\alpha_{11}, ..., \alpha_{NN})$ and \\$\boldsymbol{\beta} = \text{diag}(\beta_{11}, ..., \beta_{NN})$. Then the solution to problem \eqref{Eq:Model3} can be solved by the following message passing process
\begin{align}
    \mathbf F^{(k+1)} = \boldsymbol{\alpha}^{(k)} \mathbf{D}^{-1/2}\mathbf M^{(k)} \mathbf{D}^{-1/2} \mathbf F^{(k)} + \boldsymbol{\beta}^{(k)}\mathbf Y, \label{Eq:iter}
\end{align}
with an initial value, e.g., $\mathbf F^{(0)} = \mathbf{0}$. 
\end{theorem}

Here $\boldsymbol{\alpha}^{(k)}$, $\boldsymbol{\beta}^{(k)}$ and $\mathbf M^{(k)}$ are calculated according to the current features $\mathbf F^{(k)}$.  When $\phi(\xi) = \xi^p$, from Eq. \eqref{Eq:iter} we can have the algorithm for solving pL-UFG and pL-fUFG.
%\GaoC{Not sure whether I shall keep this remark.}
% \AHcomment{(This is very much like an attention where the the weights depend on the gradient. When $p < 2$, the aggregation assigns more weight where gradient is small and vice versa. This is also related to anisotropic diffusion.\GaoC{A reference about aniotropic diffusion here???})}
%\GaoC{The following is derivation for the algorithm. Later on we may move into Appendix.}

\textit{Proof:} 
Define $\mathcal{L}^{\phi}_p(\mathbf F)$ as the objective function in Eq. \eqref{Eq:Model3}, consider a node feature $\mathbf f_i$ in $\mathbf F$, % and define $\Delta_{i,j} =-\Delta_{j,i} = \sqrt{\frac{w_{i,j}}{d_{j,j}}}\mathbf f_j -\sqrt{\frac{w_{i,j}}{d_{i,i}}}\mathbf f_i$, 
then we have

\begin{align*}
&\frac{\partial \mathcal{L}^{\phi}_p(\mathbf F)}{\partial \mathbf{f}_i} 
=  2\mu(\mathbf f_i \!-\! \mathbf y_i)\! +\! \frac12\sum_{v_k\in\mathcal{V}}\frac{\partial}{\partial\mathbf{f}_i}\phi(\|\nabla_W\mathbf F (v_k)\|_p) \\
=&2\mu(\mathbf f_i - \mathbf y_i) + \frac12\frac{\partial}{\partial\mathbf{f}_i}\phi(\|\nabla_W\mathbf F (v_i)\|_p) 
 +\frac12\sum_{v_j\sim v_i}\frac{\partial}{\partial\mathbf{f}_i}\phi(\|\nabla_W\mathbf F (v_j)\|_p) 
\\
=&2\mu(\mathbf f_i - \mathbf y_i) + \frac12\phi'(\|\nabla_W\mathbf F (v_i)\|_p)\frac1p(\|\nabla_W\mathbf F (v_i)\|_p)^{1-p}\frac{\partial }{\partial\mathbf{f}_i}\!\left(\!\sum_{v_j\sim v_i} \|\nabla_{i,j}(w, \mathbf f)\|^p\!\right)\! \\
&+\!\frac1{2p} \sum_{v_j\sim v_i} \frac{\partial }{\partial\mathbf{f}_i}\|\nabla_{j,i}(w,\mathbf f)\|^p \cdot \phi'(\|\nabla_W\mathbf F (v_j)\|_p)(\|\nabla_W\mathbf F (v_j)\|_p)^{1-p}
 \\
=& 2\mu(\mathbf f_i - \mathbf y_i) + \frac1{2p}\phi'(\|\nabla_W\mathbf F (v_i)\|_p)(\|\nabla_W\mathbf F (v_i)\|_p)^{1-p} \cdot \\
&\quad\quad\left(\!\sum_{v_j\sim v_i}p \|\nabla_W \mathbf F ([i,j])\|^{p-2} \sqrt{\frac{w_{i,j}}{d_{i,i}}} (-\nabla_W \mathbf F ([i,j]))\!\right)\\
& + \frac12\sum_{v_j\sim v_i} \phi'(\|\nabla_W\mathbf F (v_j)\|_p) (\|\nabla_W\mathbf F (v_j)\|_p)^{1-p}  \|\nabla_W \mathbf F ([i,j])\|^{p-2} \sqrt{\frac{w_{i,j}}{d_{i,i}}}\nabla_W \mathbf F ([j,i])\\
=&2\mu(\mathbf f_i\! -\! \mathbf y_i)\! +\! \sum_{v_j\sim v_i}\! \frac12\|\nabla_W \mathbf F ([i,j])\|^{p-2}\!\!\sqrt{\frac{w_{i,j}}{d_{i,i}}}\nabla_W \mathbf F ([j,i]) \cdot \\
&\quad\quad\left[\frac{\phi'(\|\nabla_W\mathbf F (v_i)\|_p)}{\|\nabla_W\mathbf F (v_i)\|^{p-1}_p} + \frac{\phi'(\|\nabla_W\mathbf F (v_j)\|_p)}{\|\nabla_W\mathbf F (v_j)\|^{p-1}_p} \right] 
\end{align*}
Setting $\frac{\partial \mathcal{L}^{\phi}_p(\mathbf F)}{\partial \mathbf{f}_i} = 0$ gives the following first order condition:
\begin{align*}
&\bigg( \sum_{v_j\sim v_i}\frac12\left[\frac{\phi'(\|\nabla_W\mathbf F (v_i)\|_p)}{\|\nabla_W\mathbf F (v_i)\|^{p-1}_p} + \frac{\phi'(\|\nabla_W\mathbf F (v_j)\|_p)}{\|\nabla_W\mathbf F (v_j)\|^{p-1}_p} \right] \cdot \|\nabla_{i,j}(w, \mathbf f)\|^{p-2}\frac{w_{i,j}}{d_{i,i}} + 2\mu\bigg) \mathbf f_i \\
=&\sum_{v_j\sim v_i}\frac12\left[\frac{\phi'(\|\nabla_W\mathbf F (v_i)\|_p)}{\|\nabla_W\mathbf F (v_i)\|^{p-1}_p} + \frac{\phi'(\|\nabla_W\mathbf F (v_j)\|_p)}{\|\nabla_W\mathbf F (v_j)\|^{p-1}_p} \right] \cdot \|\nabla_W \mathbf F ([i,j])\|^{p-2}\frac{w_{i,j}}{\sqrt{d_{i,i}d_{j,j}}}\mathbf f_j + 2\mu \mathbf y_i.
\end{align*}
This equation defines the message passing on each node $v_i$ from its neighbour nodes $v_j$. With the definition of $M_{i,j}$, $\alpha_{ii}$ and $\beta_{ii}$, it can be turned into the iterative formula in Eq. \eqref{Eq:iter}. This completes the proof.

\begin{remark}
    When $p\leq 1$, the objective function is not differentiable in some extreme cases for example the neighbor node signals are the same and with the same degrees. In these rare cases, the first order condition cannot be applied in the above proof. However in practice, we suggest the following alternative iterative algorithm to solve the optimization problem. In fact, we can split the terms in $\mathcal{S}_p(\mathbf F)$ as
    %\EScomment{In the $p$-Lap paper, they set $M_{i,j}=0$ when considering gradient at 0, will this resolve your concern?}
\begin{align}
    \mathcal{S}_p(\mathbf F) = \frac12\sum_{(v_i,v_j)\in\mathcal{E}}\left\|\nabla_W \mathbf F ([i,j])\right\|^{p-2}\left\|\nabla_W \mathbf F ([i,j])\right\|^2.
\end{align}
At iteration $k$, we take $w^{\text{new}}_{i,j} = w_{i,j} \left\|\nabla_W \mathbf F ([i,j])\right\|^{p-2}$ as the new edge weights, then the next iterate is defined as the solution to optimize the Dirichlet energy with the new weights, i.e.,
\[
\mathbf F^{(k+1)} = \argmin %S^{\text{new}}_p(\mathbf F) = 
\frac12 \sum_{(v_i,v_j)\in\mathcal{E}} \left\|\nabla_{W^{\text{new}}} \mathbf F ([i,j])\right\|^2.
\]
Thus one step of the classic GCN can be applied in the iteration to solve the $p$-Laplacian regularized problem \eqref{Eq:2}. 
\end{remark} %\GaoC{Later on, we can use this algorithm for the case of $p=1$.}

\subsection{Interaction between $p$-Laplacian and Framelets}
In this section, we present some theoretical support on how the $p$-Laplacian regularizer interact with the framelets in the model. Specifically, we show that the proposed algorithm in Eq. \eqref{Eq:iter} provides a sequence of ($p$-Laplacian-based) spectral graph convolutions and diagonal scaling of the node features over the framelet filtered graph signals.This is indicated by the following analysis.

% \GaoC{Now we need remove Lemma 1. Instead we use a compromised argument that $F^{(R)}$ is the result of a sequence of $\Delta_p$ convolutions and diagonal multiplication layers. I am trying to organize something here based on our discussion yesterday.}
First considering the iteration Eq. \eqref{Eq:iter} with initial values $\mathbf F^{(0)} = \mathbf Y = \mathcal{W}^T\textrm{diag}(\theta)\mathcal{W}\mathbf X$ or $\textrm{diag}(\theta_{r,j}) \mathcal{W}_{r,j}  \mathbf X$ without loss of generality\footnote{In our practical algorithm we choose 
$\mathbf F^{(0)} = 0$, this will gives $\mathbf F^{(1)} = \beta^{(0)}\mathbf Y$ with a diagonal matrix $\beta^{(0)}$.}, we have 
\begin{align}
    &\mathbf F^{(K)} = \boldsymbol{\alpha}^{(K\!-\!1)} \mathbf{D}^{-1/2}\mathbf M^{(K-1)} \mathbf{D}^{-1/2} \mathbf F^{(K\!-\!1)}\! + \!\boldsymbol{\beta}^{(K-1)}\mathbf Y \notag \\
     = & \boldsymbol{\alpha}^{(K-1)}\widetilde{\mathbf M}^{(K-1)}\mathbf F^{(K-1)} + \boldsymbol{\beta}^{(K-1)}\mathbf Y\notag  \\
     = & \boldsymbol{\alpha}^{(K-1)}\widetilde{\mathbf M}^{(K-1)}\left(\boldsymbol{\alpha}^{(K-2)}\widetilde{\mathbf M}^{(K-2)}\mathbf F^{(K-2)} + \boldsymbol{\beta}^{(K-2)}\mathbf Y \right)\notag  +  \boldsymbol{\beta}^{(K-1)}\mathbf Y \notag \\
     = &\boldsymbol{\alpha}^{(K-1)}\widetilde{\mathbf M}^{(K-1)}\boldsymbol{\alpha}^{(K-2)} \widetilde{\mathbf M}^{(K-2)}\mathbf F^{(K-2)} \notag + \boldsymbol{\alpha}^{(K-1)}\widetilde{\mathbf M}^{(K-1)} \boldsymbol{\beta}^{(K-2)}\mathbf Y +\boldsymbol{\beta}^{(K-1)}\mathbf Y \notag  \\
    %&= \left(\prod_{k=0}^{K-1} \boldsymbol{\alpha}^{(k)}\right)\left(\prod_{k=0}^{K-1} \widetilde{\mathbf M}^{(k)}\right)\mathbf F^{(0)} + \boldsymbol{\beta}^{(K-1)}\mathbf Y \\
    %&+ \sum_{k=0}^{K-1} \left(\prod_{l =K-k}^{K-1} \boldsymbol{\alpha}^{(l)}\widetilde{\mathbf M}^{(l)}\right)\boldsymbol{\beta}^{(K-k-1)}\mathbf Y \\
     = &\left(\prod_{k=0}^{K-1} \boldsymbol{\alpha}^{(k)}\widetilde{\mathbf M}^{(k)}\right)\mathbf Y + \boldsymbol{\beta}^{(K-1)}\mathbf Y \notag  + \sum_{k=0}^{K-1} \left(\prod_{l =K-k}^{K-1} \boldsymbol{\alpha}^{(l)}\widetilde{\mathbf M}^{(l)}\right)\boldsymbol{\beta}^{(K-k-1)}\mathbf Y.  \label{iteration_of_F(k)}
\end{align}
The result $\mathbf F^{(K)}$ depends on the key operations $\boldsymbol{\alpha}^{(k)}\widetilde{\mathbf M}^{(k)}$ for $k = 0, 1, ..., K-1$, where 
\begin{align}
&\widetilde{M}^{(k)}_{i,j} =  \frac{w_{i,j}}{\sqrt{d_{i,i}d_{j,j}}}  \left\| \nabla_W\mathbf F^{(k)}[(i,j)] \right\|^{p-2} \!\!\! \cdot \notag \left[\frac{\phi'(\|\nabla_W\mathbf F^{(k)} (v_i)\|_p)}{\|\nabla_W\mathbf F^{(k)} (v_i)\|^{p-1}_p} + \frac{\phi'(\|\nabla_W\mathbf F^{(k)} (v_j)\|_p)}{\|\nabla_W\mathbf F^{(k)} (v_j)\|^{p-1}_p} \right] \notag \\
&=\frac{w_{i,j} \cdot w^{(k)}_{i,j}(p)}{\sqrt{d_{i,i}d_{j,j}}}  \left\| \nabla_W \mathbf F^{(k)}[(i,j)] \right\|^{p-2},  \\ 
&\alpha_{i,i}^{(k)}  =  1/\left(\sum_{v_j\sim v_i}{\widetilde{M}^{(k)}_{i,j}}  + 2\mu\right), \label{alpha_value}
\end{align}
where $\widetilde{M}_{ij}$ has absorbed $d_{i,i}$ and $d_{j,j}$ into $M_{ij}$ defined in Theorem \ref{algorithm_thm}, i.e., $\widetilde{M}_{ij} = M_{i,j}/\sqrt{d_{i,i}d_{j,j}}$. 

Denote  
$$w^{(k)}_{i,j}(p)=\left[\frac{\phi'(\|\nabla_W\mathbf F^{(k)} (v_i)\|_p)}{\|\nabla_W\mathbf F^{(k)} (v_i)\|^{p-1}_p} 
  +  \frac{\phi'(\|\nabla_W\mathbf F^{(k)} (v_j)\|_p)}{\|\nabla_W\mathbf F^{(k)} (\!v_j\!)\|^{p-1}_p} \right].$$ %$w^{(k)}_{i,j}(p) =\frac{\left[\frac{\phi'(\|\nabla_W\mathbf F^{(k)} (v_i)\|_p)}{\|\nabla_W\mathbf F^{(k)} (v_i)\|^{$p$-1}_p} + \frac{\phi'(\|\nabla_W\mathbf F^{(k)} (v_j)\|_p)}{\|\nabla_W\mathbf F^{(k)} (v_j)\|^{$p$-1}_p} \right]}{\sqrt{d_{i,i}d_{j,j}}}$. 
To introduce the following analysis, define the relevant matrices $\widetilde{\mathbf M}^{(k)} = [\widetilde{M}^{(k)}_{i,j}]$, $\boldsymbol{\alpha}^{(k)} = \text{diag}(\alpha_{ii}^{(k)} )$ and $\mathbf W^{(k)}_p = [w^{(k)}_{i,j}(p)]= \mathbf w^{(k)}_p \oplus (\mathbf {w}^{(k)}_p)^T$ (the Kronecker sum) with the column vector
 \[
 \mathbf w^{(k)}_p = \left[\frac{\phi'(\|\nabla_W\mathbf F^{(k)} (v_i)\|_p)}{\|\nabla_W\mathbf F^{(k)} (v_i)\|^{p-1}_p}\right]^N_{i=1}.
 %\mathbf w^{(k)}_p = \left[\frac{\phi'(\|\nabla_W\mathbf F^{(k)} (v_i)\|_p)}{\sqrt{d_{i,i}}\|\nabla_W\mathbf F^{(k)} (v_i)\|^{$p$-1}_p}\right]^N_{i=1}.
 \]

Now, recall that the definition of the classic $p$-Laplacian is: 
\begin{align}
    \Delta_p (\mathbf F): = - \frac{1}{2} \text{div}( \|\nabla \mathbf F\|^{p-2} \nabla \mathbf F), \quad \text{for} \,\,\, p\geq 1, \label{Eq:$p$-Laplacian}
\end{align}
Our purpose is to show that our generalized algorithm Eq. \eqref{iteration_of_F(k)} above can be implemented as the linear combination of the classic $p$-Laplacian filtering. First, based on \cite{FuZhaoBian2022}, the operator Eq. \eqref{Eq:$p$-Laplacian} in the original $p$-Laplacian message passing algorithm can be written in detailed matrix form as follows
\begin{align}
    \Delta^{(k)}_p(\mathbf F^{(k)})  = \left((\boldsymbol{\alpha}^{(k)}_0)^{-1} - 2\mu \mathbf I_N \right)\mathbf F^{(k)} -  {\mathbf G^{(k)}}        \mathbf F^{(k)}, \label{Eq:$p$-Laplacian-2}
\end{align}
where the matrix $\mathbf G^{(k)}$ has elements 
\begin{align*}
G^{(k)}_{i,j} = & \frac{w_{i,j}}{\sqrt{d_{i,i}d_{j,j}}}  \left\| \nabla_W \mathbf F^{(k)}[(i,j)] \right\|^{p-2},      
\end{align*}
and the diagonal matrix $\boldsymbol{\alpha}^{(k)}_0$ has diagonal element defined as
\begin{align*}
(\alpha^{(k)}_0)_{ii}  = & 1/\left(\sum_{v_j\sim v_i}{G^{(k)}_{i,j}} + 2\mu/p\right).
\end{align*}
Eq. \eqref{Eq:$p$-Laplacian-2} shows that the operation  $\Delta^{(k)}_p(\mathbf F^{(k)})$ can be represented as the product of a matrix (still denoted as $\Delta^{(k)}_p$) and the signal matrix $\mathbf F^{(k)}$.

Noting that $\mathbf G^{(k)} = \widetilde{\mathbf M}^{(k)}\oslash \mathbf W^{(k)}_p$, multiplying diagonal matrix $\boldsymbol{\alpha}^{(k)}$ on both sides of Eq. \eqref{Eq:$p$-Laplacian-2} and taking out $\mathbf F^{(k)}$ give
\begin{align*}
   \boldsymbol{\alpha}^{(k)} \Delta^{(k)}_p   = \left(\boldsymbol{\alpha}^{(k)}(\boldsymbol{\alpha}^{(k)}_0)^{-1} - 2\mu \boldsymbol{\alpha}^{(k)} \right) - \boldsymbol{\alpha}^{(k)} (\widetilde{\mathbf M}^{(k)}\oslash \mathbf W^{(k)}_p).        %\label{Eq:$p$-Laplacian-3}
\end{align*}
As $\boldsymbol{\alpha}^{(k)}$ is diagonal, we can re-write the above equality as 
\begin{align*}
\boldsymbol{\alpha}^{(k)} \widetilde{\mathbf M}^{(k)}  =  \left(\left(\boldsymbol{\alpha}^{(k)}(\boldsymbol{\alpha}^{(k)}_0)^{-1} -  2\mu \boldsymbol{\alpha}^{(k)} \right)  - \boldsymbol{\alpha}^{(k)} \Delta^{(k)}_p  \right) \odot \mathbf W^{(k)}_p.        %\label{Eq:$p$-Laplacian-3}
\end{align*}

Given that $\mathbf W^{(k)}_p$ is the Kronecker sum of the vector $\mathbf w^{(k)}_p$ and its transpose, if we still use  $\mathbf w^{(k)}_p$ as its diagonal matrix, then we have
\begin{align}\label{filtering_interaction}
\boldsymbol{\alpha}^{(k)} \widetilde{\mathbf M}^{(k)}  = &  \left(\left(\boldsymbol{\alpha}^{(k)}/\boldsymbol{\alpha}^{(k)}_0 - 2\mu \boldsymbol{\alpha}^{(k)} \right)  -  \boldsymbol{\alpha}^{(k)} \Delta^{(k)}_p \right)   \mathbf w^{(k)}_p \notag \\
 + \mathbf w^{(k)}_p  &\left(\left(\boldsymbol{\alpha}^{(k)}/\boldsymbol{\alpha}^{(k)}_0 - 2\mu \boldsymbol{\alpha}^{(k)} \right)  -  \boldsymbol{\alpha}^{(k)} \Delta^{(k)}_p \right). 
\end{align}

This demonstrates that the key terms $\boldsymbol{\alpha}^{(k)} \widetilde{\mathbf M}^{(k)}$ in our generalized algorithm is in linear form of $p$-Laplacian operator $\Delta^{(k)}_p$. As demonstrated in the research \cite{FuZhaoBian2022}, the operator $\Delta^{(k)}_p$ approximately performs spectral graph convolution. Hence we can conclude that the generalized $p$-Laplacian iterations \eqref{Eq:iter}  indeed performs a sequence of graph spectral convolutions (see Lemma \ref{lemma1} below) and gradient-based diagonal transformation (i.e., node feature scaling) over the framelet filtered graph signals. 

\begin{lem}\label{lemma1}The matrix $\Delta^{(k)}_p$ is SPD (see \cite{bozorgnia2019first,ly2005first}) which has its own eigendecomposition, offering a graph spectral convolution interpretation.  %, \GaoC{Ethan please add your references here.}
\end{lem}

\begin{remark}
Eq. \eqref{filtering_interaction} indicates that the $p$-Laplacian regularizer provides graph spectral convolution on top of the framelet filtering, which produces a second layer of filtering conceptually and hence restricts the solution space further. See Fig. \ref{fig:regu} for the illustration. Interestingly, the combination of $p$-Laplacian regularization and  framelet offers a more adaptive smoothness that suites both homophilic and heterophilic data as shown in our experiments. 
%further restricts the solution space of original framelet with different degree of smoothness, as we have shown in in Fig. \ref{fig:regu}. Furthermore, as we involve a more general convex function $\phi$ to penalize the variation of signal $\mathbf F$, it is natural for us to incorporate $\phi$ with a flexible (i.e., multi-scale) GNN model like framelet. This shows the effectiveness of the framelet model in terms of adapting general GNN regularization problem. 
% In equation \eqref{filtering_interaction} the algorithm presented in  \eqref{Eq:iter} shows a second filtering operation onto the framelet decomposed signal $\mathbf Y$. This aligns with the spirit that by assigning a different scale of penalty term (i.e., $p$-Laplacian regularizer) into the GNN regularization problem, we restrict the solution space of the model's outputs, as we have shown in Fig. \ref{fig:regu}. 
\end{remark}

\begin{remark}\label{asymptotic_behavior}
In terms of asymptotic behavior of Eq. \eqref{iteration_of_F(k)}, one can show roughly that the elements in $\boldsymbol{\alpha}^{(k)}\widetilde{\mathbf M}^{(k)}$ are between 0 and 1, which converges to zero if $K$ is too large. Therefore a large $K$ annihilates the first term in Eq. \eqref{iteration_of_F(k)} but leaves the second term and partial sum from the third. A larger value of $\mu$ seems to speed up this convergence further, resulting shortening the time for finding the solution. However, it is a model selection problem for generalizability. Moreover, the inclusion of the source term $\boldsymbol{\beta}^{(K-1)}\mathbf Y$ guarantees to supply certain amount of variations of the node features from both low and high frequency domains of the graph framelet and it has been shown such inclusion can help the model escape from over-smoothing issue \citep{chien2020adaptive}.
%In addition to the asymptotic behavior of equation \eqref{iteration_of_F(k)}, we note that based on the form of $\alpha$ illustrated in equation \eqref{alpha_value}, we have $0<\alpha_{i,i}<1$. Hence $\alpha$ provides a diluting effect on its attached $\mathbf Y$, in fact when $K \rightarrow \infty$, $\prod_{k=0}^{K-1} \boldsymbol{\alpha}^{(k)} \rightarrow 0$, inducing a significant model diluting mechanism to the model. 
%Due to this case, we may require a relatively large value of $\mu$ in the term of $\boldsymbol{\beta}^{(K-1)}$ by not only maintaining a certain amount of information from the source but also making the regularization problem (i.e., equation \eqref{Eq:1}) further dominated by the framelet model.This conclusion is further confirmed by comprehensive empirical studies see section \ref{Sec:5} for more details.
%\EScomment{need to be double checked.}
% \GaoC{Unfinished??}
\end{remark}

\begin{remark}
Compared to GPRGNN \citep{chien2020adaptive} in which the model outcome is obtained by 
\begin{align*}
\hat{\mathbf F}^{(k)} = \sum_k^{K-1}\boldsymbol{\gamma}_k\mathbf F^{(k)},
\end{align*}
where $\boldsymbol{\gamma}_k \in \mathbb R^{N\times N}$ is learnable generalized Page rank coefficients,
as we have shown in Eq. \eqref{filtering_interaction}, our proposed algorithm defined in Eq. \eqref{Eq:iter} provides a mixed (spectral convolution and diagonal scaling) to the framelet graph signal outputs and thus further directs the solution space of framelet to a reasonably defined region.  Furthermore, due to the utilization of the Chebyshev polynomial in framelet, the computational complexity for the first filtering is not high, which is helpful in terms of defining the provably corrected space for the second operation. This shows the efficiency of incorporating framelet in $p$-Laplacian based regularization.
\end{remark}

\EScomment{
\section{Discussions on the Proposed Model}
In this section, we conduct comprehensive discussions for our proposed models. We note that we will mainly focused on exploring the property of the generalized $p$-Laplacian based framelet GNN (Eq.~\eqref{Eq:Model3}) and the iterative algorithm in Eq.~\eqref{Eq:iter}, although we may also show some conclusions of pL-UFG and pL-fUFG as well. Specifically, in Section \ref{discussion_denoising_power} we discuss the denoising power of our proposed model. Section \ref{computational_complexity} analyzes the model's computational complexity. Section \ref{Regularizer_comparison} provides a comparison between the $p$-Laplacian regularizer with other regularizers that are applied to GNNs via different learning tasks. Finally, the section is concluded (Section \ref{limitation_and_future}) by illustrating the limitation of the model and potential aspects for future studies.}

\subsection{Discussion on the Denoising Power}\label{discussion_denoising_power}
%\AHcomment{Move the denoising section here}
The denoising power of framelet has been developed in the study\citep{Dong2017} and empirically studied by the past works \cite{Dong2017,ZhouLiZhengWangGao2021,yang2022quasi}. Let $\mathbf X = \mathbf F+ \mathbf \tau$ be the input node feature matrix with noise $\mathbf \tau$. In works \citep{Dong2017,ZhouLiZhengWangGao2021}, a framelet-based regularizer was proposed to resolve the optimization problem which can be described as: 
\begin{align}\label{denoise_initial}
    \min_\mathbf{F} \|\mathcal D\mathbf F\|_{1,\mathcal G}+ f(\mathbf F, \mathbf X),
\end{align}
where $f(\cdot,\cdot)$ is some fidelity function that takes different forms for different applications. For any graph signal $\mathbf F$, the graph $\mathbb L_p$-norm $\|\mathbf F\|_{p,\mathcal G} := \left(\sum_i \vert \mathbf f_i\vert^p \times d_{i,i}\right)^{1/p}$, where $d_{i,i}$ is the degree of the $i$-th node with respect to $\mathbf f$. $\mathcal D $ is a linear transformation generated from discrete transformations, for example, the graph Fourier transforms or the Framelet transforms. Thus, the regularizer assigns a higher penalty to the node with a larger number of neighbours. Specifically,  considering the functional $\mathcal D$ as the framelet transformation, then the first term in Eq. \eqref{denoise_initial} can be written as: 

\begin{align}\label{framelet_regularizer}
    \|\mathcal D\mathbf F\|_{1,\mathcal G} = \left( \sum_{(r,\ell)\in\mathcal Z}\sum_i \vert \mathbf F_{r,\ell}[i]\vert^p \times d_{i,i}\right)^{1/p}.
\end{align}

where $\mathcal Z = \{ (r,\ell) : r = 1,...,R, \ell= 0,...,J \} \cup \{ (0,J)\}$. 
Replacing the $p$-Laplacian regularizer in Eq. \eqref{Eq:1} by the framelet regularizer defined in the above Eq. \eqref{framelet_regularizer} we have: 

\begin{align}\label{framelet_deno}
    \mathbf F = \argmin_{\mathbf{F}} & \left( \sum_{(r,\ell)\in\mathcal Z}\sum_i \vert \mathbf F_{r,\ell}[i]\vert^p \times d_{i,i} \right)^{1/p} +\mu \|\mathbf{F} - \mathcal{W}^T\textrm{diag}(\theta)\mathcal{W}\mathbf X  \|^2_F. 
\end{align}

Followed by the work in the research \citep{Dong2017}, the framelet regularizer-based denoising problem in Eq. \eqref{denoise_initial} is associated with the variational regularization problem that can be generally described in the form similar to Eq. \eqref{Eq:1} where the variational term $\mathcal S_p(\mathbf F)$ is utilized for regularizing the framelet objective function. Simply by replacing the input of the first term in Eq. \eqref{framelet_deno} by $\nabla \mathbf{F}$ and omitting node degree information, we have: 
\begin{equation}
\mathbf F = \argmin_{\mathbf{F}}  \sum_{(r, \ell)\in\mathcal Z}\|\nabla \mathbf F_{r,\ell}\|_p^p+\mu \|\mathbf{F} -\mathcal{W}^T\textrm{diag}(\theta)\mathcal{W}\mathbf X\|^2_F.    
\end{equation} 

% \begin{align}
%     \mathbf F &\!=\! \argmin_{\mathbf{F}} \! \sum_{\mathcal Z}\|\nabla \mathbf F_\mathcal Z\|_p^p+\!\mu \|\mathbf{F} \!-\! \mathcal{W}^T\!\textrm{diag}(\!\theta\!)\mathcal{W}\mathbf X\|^2_F \notag\\
%     &=\sum_{v_i\in\mathcal{V}} \phi(\|\nabla_W\mathbf F (v_i)\|_p) +\mu \|\mathbf{F} \!-\! \mathcal{W}^T\!\textrm{diag}(\!\theta\!)\mathcal{W}\mathbf X\|^2_F. 
% \end{align}
%Since we have $\phi(\xi) = \xi^p$, 
Therefore our proposed model naturally is equipped with denoising capability and the corresponding (denoising) regularizing term aligns with the denoising regularizer developed in the paper \citep{ZhouSchoelkopf2005} without nodes degree information. However, the work in the study \citep{ZhouSchoelkopf2005} only handles $p=1$ and $p=2$. Whereas our $p$-Laplacian framelet model covers the values of $p \in \mathbb R_{+}$. This allows more flexibility and effectiveness in the model. %of our denoising regularizer. 

Furthermore, the major difference between most wavelet frame models and the classic variational models is the choice of the underlying transformation (i.e., $\mathcal D$ applied to the graph signal) that maps the data to the transformed domain that can be sparsely approximated. Given both $p$-Laplacian (i.e., Eq. \eqref{Eq:1}) and framelet regularizers (i.e., Eq. \eqref{framelet_regularizer}) target on the graph signal $\mathbf F_\mathcal Z$ that is produced from the sparse and tight framelet transformation (Chebyshev polynomials), this further illustrates the equivalence between two denoising regularizers. Please refer to \citep{Dong2017} for more details. 
% \EScomment{I think the denoising section is fine at lest in terms of comparing itself with the framelet denoising regularizer. Just wondering do we need some further analysis on this?}
% \subsection{Theoretical Analysis on HFD}
\EScomment{
\subsection{Discussion on the Computational Complexity}\label{computational_complexity}
In this section, we briefly discuss the computational complexity of our proposed model, specific to the generalized model defined in Eq.~\eqref{Eq:Model3} with its results that can be approximated by the algorithm presented in Eq.~\eqref{Eq:iter}. The primary computational cost of our model stems from the generation of the framelet decomposition matrices ($\mathcal W$) or the framelet transformation applied to the node features. We acknowledge that generating $\mathcal W$ involves a high computational complexity, as transitional framelet methods typically employ eigen-decomposition of the graph Laplacian for this purpose. Consequently, the framelet transform itself exhibits a time complexity of $\mathcal{O}(N^2(nj+1)Kc)$ and a space complexity of $\mathcal{O}(N^2(nj+1)c)$ for a graph with $N$ nodes and $c$-dimensional features. Notably, the constants $n$, $j$, and $K$ are independent of the graph data. Existing literature \citep{ZhengZhouGaoWangLioLiMontufar2021} has demonstrated that when the input graph contains fewer than $4\times 10^4$ nodes (with fixed feature dimension), the computational time for framelet convolution is comparable to that of graph attention networks with 8 attention heads \citep{velivckovic2017graph}. As the graph's node size increases, the performance of graph attention networks degrades rapidly, while graph framelets maintain faster computation. However, our application of Chebyshev polynomials to approximate $\mathcal W$ significantly reduces the associated computational cost compared to traditional methods. Additionally, we acknowledge that the inclusion of the $p$-Laplacian based implicit layer introduces additional computational cost to the original framelet model, primarily arising from the computation of the norm of the graph gradient, denoted as $|\nabla \mathbf F|$. Considering the example of the Euclidean norm, the computational cost for $|\nabla \mathbf F|$ scales as $\mathcal O(N\times c)$, where $N$ and $c$ represent the number of nodes and feature dimension, respectively. Thus, even when accounting for the computation of the implicit layer, the overall cost of our proposed method remains comparable to that of the framelet model.
}

\EScomment{
\subsection{Comparison with Other Regularizers and Potential Application Scenarios}\label{Regularizer_comparison}
As we have illustrated in Section \ref{More General Regularization}, with the assistance from different form of the regularizers, GNNs performance could be enhanced via different learning tasks. 
\EScomment{In this discussion, we position our study within the wider research landscape that investigates various regularizers to enhance the performance of Graph Neural Networks (GNNs) across different learning tasks. We earlier discussed the denoising power of the pL-UFG in Section \ref{discussion_denoising_power}, establishing that it can be expressed in terms of a denoising formulation. This is comparable to the approach from \cite{MaLiuZhao2021}, who used a different regularizer term to highlight the denoising capabilities of GNNs. They showed that their regularizer can effectively reduce noise and enhance the robustness of GNN models.

Our study, however, emphasizes the unique advantages that the $p$-Laplacian brings, a theme also echoed in the works by \cite{liu2010robust11} and \cite{candes2011robust12}. Both studies incorporated the $L_1$-norm as a regularization term in robust principal component analysis (RPCA), showcasing its ability to recover low-rank matrices even in the presence of significant noise. Furthermore, the study by \cite{liu2021elastic13} reinforces the benefits of the $L_1$-norm in preserving discontinuity and enhancing local smoothness. Characterized by its heavy-tail property, the $L_1$-norm imposes less penalty on large values, thereby making it effective in handling data with substantial corruption or outliers.

Furthermore, in terms of the potential practical implementations, one can deploy our method into various aspects. For example, the $p$-Laplacian based regularizer can be used in image processing and computer vision applications, where it can help to smooth out noisy or jagged images while preserving important features. In addition, we note that, as we implement our graph learning method via an optimization (regularization) framework, this suggests any potential practical implementations (such as graph generation \citep{ho2020denoising}, graph based time series forecasting \citep{wen2023diffstg}) of GNNs can also be deployed under our proposed methods with a higher flexibility power. % Given the above, it is not hard to check that our approach utilizing the $p$-Laplacian not only aligns with existing research but also introduces significant enhancements. This supports the continued exploration and refinement of such methods in future studies.
}

\subsection{Limitation and Potential Future Studies}\label{limitation_and_future}
While outstanding properties have been presented from our proposed model, the exploration of how the $p$-Laplacian based regularization framework can be deployed to various types of graphs (i.e., dynamic or spatio-temporal) is still wanted. In fact, one may require the corresponding GNNs to be able to capture the pattern of the evolution (i.e., combing GNN with LSTM or Transformer \citep{manessi2020dynamic}) of the graph when the input graph is dynamic. We note that such a requirement is beyond the capability of the current model. However, as the $p$-Laplacian based regularizer restricts the solution of the graph framelet, it would be interesting to make the quantity of $p$ learnable throughout the training process. We leave this to future work. 

Moreover, followed by the idea of making $p$ value as a learnable parameter in the model, one can also explore assigning different values of $p$ to different frequency domains of framelet. It has been verified that the low-frequency domain of framelet induces a smoothing effect on the graph signal whereas the sharpening effect is produced from the high-frequency domain \citep{han2022generalizedenergy}. Therefore, one may prefer to obtain a relatively large quantity of $p$ to the low-frequency domain to enhance the smoothing effect of the framelet when the input graph is highly homophily, as one prefers to predict identical labels for connected nodes. On the other hand, a smaller value of $p$ is wanted for a high-frequency domain to further sharpen the differences between node features (i.e., $\nabla \mathbf F$) so that distinguished labels can be produced from the model when the input graph is heterophilic. Moreover, as the $p$-Laplacian-based implicit layers are allocated before the framelet reconstruction, it would be interesting to explore how such implicit layers can affect the signal reconstruction of the framelet.
}

\section{Experiments}\label{Sec:5}
\EScomment{
In this section, we present empirical studies on pL-UFG and pL-fUFG on real-world node classification tasks with both heterophilic and homophilic graphs. We also test the robustness of the proposed models against noise. Both two experiments are presented with detailed discussions on their results. In addition, we discuss by adjusting the quantity of the $p$ in our proposed model, the so-called ablation study is automatically conducted in our experiments. The code for our experiment can be accessed via \url{https://github.com/superca729/pL-UFG}.
Lastly, it is worth noting that our proposed method has the potential to be applied to the graph learning task other than node classification such as graph level classification (pooling) \citep{ZhengZhouGaoWangLioLiMontufar2021} and link prediction \citep{LinGao2022}. Although we have yet to delve into these tasks, we believe that by assigning some simple manipulations to our methods, such as deploying the readout function for graph pooling or computing the log-likelihood for graph link prediction, our method is capable of handling these tasks.  Similarly, our model could be beneficial in community detection tasks by possibly identifying clusters of similar characteristics or behaviors. We leave these promising research aspects to our future work.}

\subsection{Datasets, Baseline Models and the Parameter Setting} \label{ssec:num1}
\indent\textbf{Datasets.} %To compare our results with the other baseline models, 
We use both homophilic and heterophilic graphs from \url{https://www.pyg.org/} to assess pL-UFG and pL-fUFG, including %These datasets can be obtained from \texttt{torch geometric.datasets}: 
benchmark heterophilic datasets: %webpage graphs 
\texttt{Chameleon}, \texttt{Squirrel}, \texttt{Actor}, \texttt{Wisconsin}, \texttt{Texas}, \texttt{Cornell}, % \citep{PeiWeiChang2020}, 
and homophilic datasets: \texttt{Cora}, \texttt{CiteSeer}, \texttt{PubMed}, \texttt{Computers},  \texttt{Photos}, \texttt{CS}, \texttt{Physics}. % \citep{SenNamataBilgicGetoorGalligherEliassiRad2008}. 
In our experiments, homophilic graphs are undirected and heterophilic graphs are directed (where we observe an improved performance when direction information is provided).
We included the summary statistics of the included datasets together with their homophily index and split ratio in Table \ref{tab:my_label1}
% The ratio of training, validation, and testing sets are shown in Table \ref{tab:my_label1}.

\indent\textbf{Baseline models.}
We consider eight baseline models for comparison:
\begin{itemize}
\item \textbf{MLP}: standard feedward multiple layer perceptron. %MLPs are neural network models that work as universal approximators,  they are composed of neurons called perceptions. %\GaoC{Say something}  
\item \textbf{GCN} \citep{KipfWelling2016}: GCN is the first of its kind to implement linear approximation to spectral graph convolutions.
\item \textbf{SGC} \citep{WuZhangSouza2019}: SGC reduces GCNs’ complexity by removing nonlinearities and collapsing weight matrices
between consecutive layers. %This method can also be regarded as decoupling the feature transformation and propagation.\\
\item \textbf{GAT} \citep{velivckovic2017graph}: GAT is a graph neural network that applies
the attention mechanism on node feature to learn edge weight.
\item \textbf{JKNet}\ \citep{XuLiTianSonobe2018}: JKNet can flexibly leverage different neighbourhood ranges to enable better structure-aware representation for each
node. %\GaoC{Say something} 
\item \textbf{APPNP} \citep{GasteigerBojchevskiGunnemann2018}: APPNP combines GNN with personalized PageRank to separate the neural network from the propagation scheme.
\item \textbf{GPRGNN} \citep{chien2020adaptive}: GPRGNN architecture that adaptively learns the GPR (General Pagerank) weights so as to jointly optimize node feature and topological information extraction, regardless the level of homophily on a graph. 
\item \textbf{$p$-GNN} \citep{FuZhaoBian2022}: $p$-GNN is $p$-Laplacian based GNN model, whose message-passing mechanism is derived from a discrete regularization framework.
\item \textbf{UFG} \citep{ZhengZhouGaoWangLioLiMontufar2021}: UFG is a type of GNNs based on framelet transforms, the framelet decomposition can naturally aggregate the graph features into low-pass and high-pass spectra. 
\end{itemize}
%In order to evaluate the efficacy of pL-UFG and pL-fUFG models, we also utilise UFG \citep{ZhengZhouGaoWangLioLiMontufar2021} as a baseline. 
The test results are reproduced using code that runs in our machine, which might be different from the reported results. In addition, compared to $p$-GNN, the extra time and space complexity induced from our algorithm is $O(N^2(RJ+1)d)$.

\textbf{Hyperparameter tuning.} We use grid search for tuning hyperparameters. We test $p \in\{1.0, 1.5, 2.0, 2.5\}$ for PGNN, pL-UFG and pL-fUFG. 
The learning rate is chosen from \{0.01, 0.005\}. We consider the number of iterations $T$ in $p$-Laplacian message passing from $\{4, 5\}$ after 10 warming-up steps. For homophilic datasets, we tune $\mu \in \{0.1, 0.5, 1, 5, 10\}$ and for heterophilic graphs $\mu \in \{3, 5, 10, 20, 30, 50, 70\}$.
The framelet type is fixed as Linear (see \citep{yang2022quasi}) and the level $J$ is set to 1. The dilation scale ${s} \in \{1, 1.5, 2, 3, 6\}$, and for $n$, the degree of Chebyshev polynomial approximation to all $g$'s in \eqref{Eq:1}-\eqref{Eq:5}, is drawn from \{2, 3, 7\}. \EScomment{It is worth noting that in graph framelets, the Chebyshev polynomial is utilized for approximating the spectral filtering of the Laplacian eigenvalues. Thus, a $d$-degree polynomial approximates $d$-hop neighbouring information of each node of the graph. Therefore, when the input graph is heterophilic, one may require $d$ to be relatively larger as node labels tend to be different between directly connected (1-hop) nodes.} The number of epochs is set to 200, the same as the baseline model\citep{FuZhaoBian2022}.

\begin{table}[t]
\label{datasets}
\centering
\caption{Statistics of the datasets, $\mathcal H(\mathcal G)$ represent the level of homophily of overall benchmark datasets.}
\label{tab:my_label1}
\setlength{\tabcolsep}{3pt}
\renewcommand{\arraystretch}{1.5}
    \begin{tabular}{ccccccc}
    \hline
         Datasets & Class & Feature & Node & Edge & Train/Valid/Test & $\mathcal H(\mathcal G)$\\
         \hline
        Cora & 7 & 1433 & 2708 & 5278 & 20\%/10\%/70\% & 0.825\\
        CiteSeer & 6 & 3703 & 3327 & 4552 & 20\%/10\%/70\%  & 0.717\\
        PubMed & 3 & 500 & 19717 & 44324 & 20\%/10\%/70\%  & 0.792\\
        Computers & 10 & 767 & 13381 & 245778 & 20\%/10\%/70\%  & 0.802\\
        Photo & 8 & 745 & 7487 & 119043 & 20\%/10\%/70\%  & 0.849\\
        CS & 15 & 6805 & 18333 & 81894 & 20\%/10\%/70\%  & 0.832\\
        Physics & 5 & 8415 & 34493 & 247962 & 20\%/10\%/70\%  & 0.915\\
        \hline
        Chameleon &5 &2325 &2277 &31371 &60\%/20\%/20\% &0.247\\
        Squirrel &5 &2089 &5201 &198353 &60\%/20\%/20\% &0.216\\
        Actor &5 &932 &7600 &26659 &60\%/20\%/20\% &0.221\\
        Wisconsin &5 &251 &499 &1703 &60\%/20\%/20\% &0.150\\
        Texas &5 &1703 &183 &279 &60\%/20\%/20\% &0.097\\
        Cornell &5 &1703 &183 &277 &60\%/20\%/20\% &0.386\\
        \hline
    \end{tabular}   
\end{table}

\subsection{Experiment Results and Discussion} 
\textbf{Node Classification} Tables \ref{tab:my_label2} and \ref{tab:my_label4} summarize the results on homophilic and heterophilic datasets. The values after $\pm$ are standard deviations. The top three results are highlighted in \textbf{\textcolor{red}{First}}, \textbf{\textcolor{blue}{Second}}, and \textbf{Third}. From the experiment results, we observe that pL-UFG and pL-fUFG achieve competitive performances against the baselines in most of the homophily and heterophily benchmarks. For the models' performance on homophily undirect graphs, Table \ref{tab:my_label2} shows that  pL-UFG2$^{1.5}$ has the top accuracy in Cora, Photos and CS. For Citeseer, pL-UFG1$^{2.0}$, pL-UFG2$^{2.0}$ and pL-fUFG$^{2.0}$ have the best performance. In terms of the performance on heterophily direct graphs (Table \ref{tab:my_label4}), we observe that both of our two models are capable of generating the highest accuracy in all benchmark datasets except Actor where the top performance was generated from MLP. However, we note that for \texttt{Actor}, our pL-UFG1$^{1.0}$ achieves almost identical outcomes compared to those of MLP.  
% \question{One reviewer requires a further discussion on why the result is not ideal for Actor, but I don't think changes are needed here?}

% , pL-UFG and pL-fUFG obtain a superior performance against most state of art GNNs on most real-world datasets except the Actor dataset. For Actor, MLP works very well and significantly outperformed other models, the result of pL-UFG1$^{1.0}$ is extremely close to the performance of MLP.

Aligned with our theoretical prediction in Remark \ref{asymptotic_behavior} from the experiment, we discover that the value of the trade-off term $\mu$ for pL-UFG and pL-fUFG is significantly higher than that in pGNN, indicating that framelet has a major impact on the performance of the model. The performance of pL-UFG, pL-fUFG and others on heterophilic graphs are shown in Table \ref{tab:my_label4}. pL-UFG and pL-fUFG both can outperform MLP and other state-of-the-art GNNs under a low homophilic rate. In terms of denoising capacity, pL-UFG and pL-fUFG are far better than the baseline models. Figs. \ref{fig:denoise_cora} and \ref{fig:denoise_chamel} show that both pL-UFG and pL-fUFG produce the top accuracies across different noise levels.
%Finally, we conducted denoising experiments compared with PGNN.  

% \EScomment{Not sure why this paragraph is necessary? }

%\subsection{Discussions}

%\subsection{Discussion on selecting p}
\textbf{Discussion for $\mathbf{p < 2}$.}
To enable GNN model to better adapt to the heterophilic graphs, an ideal GNN model shall induce a relatively high variation in terms of the node feature generated from it.
% sharpen the  nodes information rather than smoothing it out. 
Therefore, compared to the models with $p=2$, the model regularized with $p \neq 2$ regularizer imposes a lesser penalty and thus produces outputs with a higher variation. This is supported by the empirical observation from Table \ref{tab:my_label4} in which the highest prediction accuracy for heterophilic graphs are usually achieved by our model with $p <2$.

% Comparing $p < 2$ with $p = 2$ (known as Dirichlet energy regularization), we note that the case of $p < 2$ imposes a lesser penalty for large gradient $\nabla_W \mathbf F$  than $p = 2$. Thus, the local variation is preserved, ensuring better model discriminative power (particularly for heterophilic graphs). The framelets utilise the retained local variation (from $p <2$) as high-frequency node information so that the over-smoothing is avoided. 

% This also can explain the results in Table \ref{tab:my_label4}. Node classification for heterophilic graphs where the nodes have a great level of variation, benefits from the retained high-frequency information from the proposed models. 

\textbf{Discussion for $p$ = 1.}
Here we specifically focus on $p=1$ in our models. %In this paragraph, we included some further discussion when our model with $p=1$. 
Recall that when $p=1$, the $p$-Laplacian operator acts on the graph signal $\mathbf F$ is $\Delta_1 \mathbf F = -\frac12\text{div}\left(\frac{\nabla \mathbf F}{\|\nabla \mathbf F\|}\right)$. Based on 
Remark \ref{p_1_explain}, when $p=1$, $\Delta_1 \mathbf F$ is equivalent to the mean curvature operator defined on the embedded surface. % of the curve. 
In analogy to differential geometry, points that curvature can not be properly defined are so-called \textit{singular points}. Thus one can similarly conclude that when a graph contains a singular node (i.e., the graph is crumpled \citep{burda2001statistical}) then both $\Delta_1 \mathbf F$ and its corresponding $\mathcal S_p (\mathbf F)$ can not be properly defined. 
Furthermore, one can easily check that, when $p=1$, the regularization term $\phi(\xi) = \xi^{p}$ produces a higher penalty than its $\mathcal S_p(\mathbf F)$ counterparts. This is because, when $p=1$, $\mathcal S_p(\mathbf F) = (\sum_i^c (\nabla \mathbf f_i)^2)^{\frac{1}{2}}$ whereas  $\phi(\xi) = \xi^{1} = \sum_i \vert\nabla \mathbf f_i\vert$. Clearly, we have $\mathcal S_p(\mathbf F) < \phi(\xi)$ unless all of the nodes in the graph are \textit{singular nodes}, or in the graph in which there is only one non-singular point while the rest nodes are singular.

\EScomment{
\subsection{Visualization on the Effect of $p$}
Based on the claim we made in previous sections, increasing the value of $p$ leads our model to exhibit a stronger smoothing effect on the node features, thereby making it more suitable for homophilic graphs. Conversely, when $p$ is small, the model preserves distinct node features, making it a better fit for heterophilic graphs. To validate this idea, we visualize the changes in relative positions (distances) between node features for $p=1$ (the minimum value) and $p=2.5$ (the maximum value). We specifically selected the \texttt{Cora} and \texttt{Actor} datasets and employed Isomap \citep{tenenbaum2000global} to map both the initial node features and the node features generated by our model to $\mathbb R^2$, while preserving their distances. The results are depicted in Figure~\ref{isomap_homo} and \ref{isomap_hetero}.

From both Figure~\ref{isomap_homo} and \ref{isomap_hetero}, it can be observed that when $p=1$, the sharpening effect induced by our model causes the node features to become more distinct from each other. Under the Isomap projection, the nodes are scattered apart compared to the input features. Conversely, when a relatively large value of $p=2.5$ is assigned, the model exhibits a stronger smoothing effect, resulting in all nodes being aggregated towards the center in the Isomap visualization. These observations provide direct support for our main claim regarding the proposed model.

% \begin{figure}[H]
%      \centering
%      \begin{subfigure}[b]{0.01\textwidth}
%          \centering
%          \includegraphics[width=\textwidth]{actor_p1_epoch1.png}
%          \caption{Original input}
%      \end{subfigure}
%      \hfill
%      \begin{subfigure}[b]{0.1\textwidth}
%          \centering
%          \includegraphics[width=\textwidth]{actor_p1_epoch150.png}
%          \caption{$p=1$ at epoch 200}
%      \end{subfigure}
%      \hfill
%      \begin{subfigure}[b]{0.1\textwidth}
%          \centering
%          \includegraphics[width=\textwidth]{actor_p2.5_epoch150.png}
%          \caption{$p=2.5$ at epoch 200}
%      \end{subfigure}
%         \caption{Isomap Visualization with using heterophilic data (\texttt{Actor}) by changing of $p$.}
%         \label{isomap_hetero}
% \end{figure}

% \begin{figure}[H]
%      \centering
%      \begin{subfigure}[b]{0.1\textwidth}
%          \centering
%          \includegraphics[width=\textwidth]{cora_p1_epoch1.png}
%          \caption{Initial}
%      \end{subfigure}
%      \hfill
%      \begin{subfigure}[b]{0.1\textwidth}
%          \centering
%          \includegraphics[width=\textwidth]{cora_p1_epoch150.png}
%          \caption{$p=1$ at epoch 200}
%      \end{subfigure}
%      \hfill
%      \begin{subfigure}[b]{0.1\textwidth}
%          \centering
%          \includegraphics[width=\textwidth]{cora_p2.5_epoch150.png}
%          \caption{$p=2.5$ at epoch 200}
%      \end{subfigure}
%         \caption{Isomap Visualization with using homophilic data(\texttt{Cora}) by changing of $p$.}
%         \label{isomap_homo}
% \end{figure}

\begin{figure}[t]
    \centering
    \includegraphics[scale = 0.24]{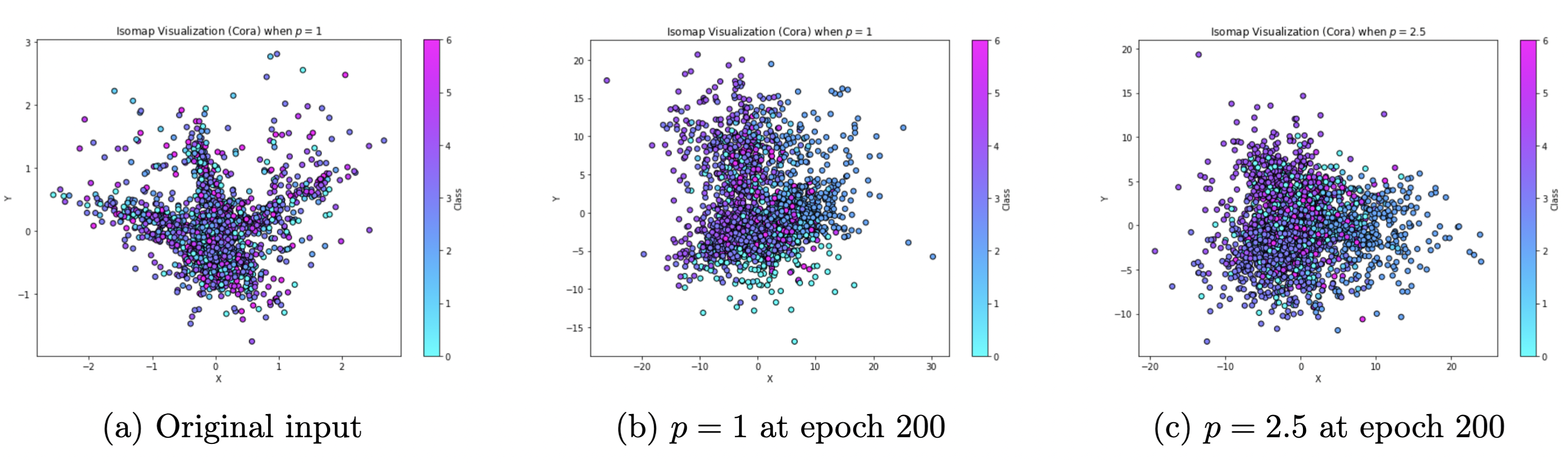}
    \caption{Isomap Visualization with using homophilic data (\texttt{Cora}) by changing of $p$.}
    \label{isomap_homo}
\end{figure}

\begin{figure}[t]
    \centering
    \includegraphics[scale = 0.27]{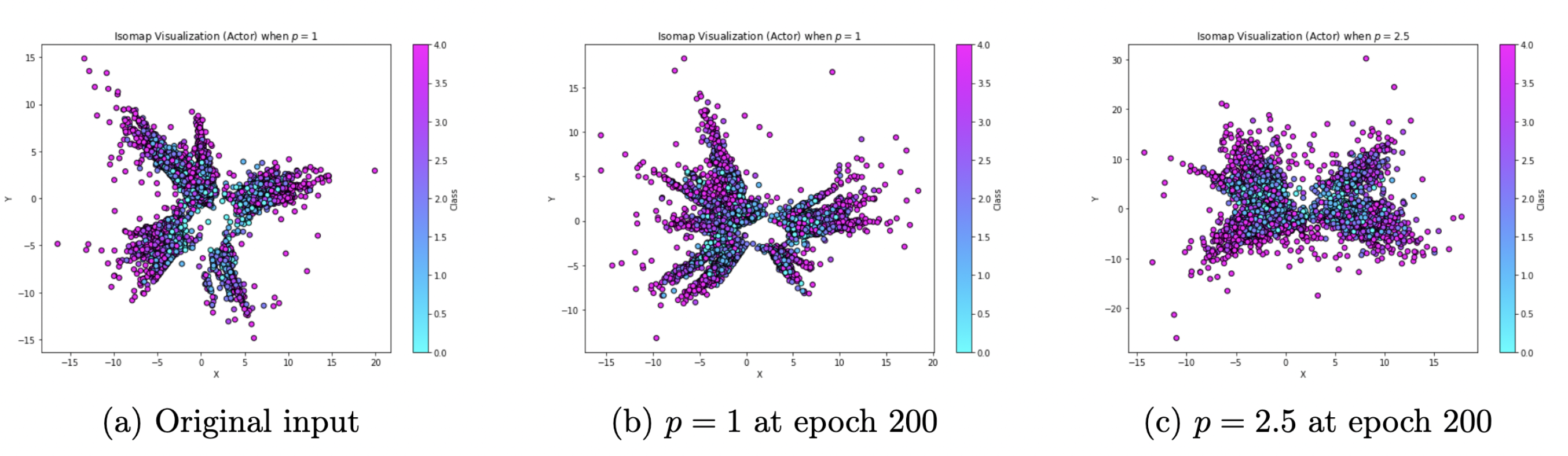}
    \caption{Isomap Visualization with using heterophilic data (\texttt{Actor}) by changing of $p$.}
    \label{isomap_hetero}
\end{figure}

}

\begin{table*}[t]
\setlength{\tabcolsep}{1pt}
\renewcommand{\arraystretch}{1.5}
\caption{Test accuracy (\%) on homophilic undirected graph.} %{\color{red}I remember in their paper, they used 2.5\% for training and 2.5\% for validation??}}{\color{orange} I tested on 0.025 0.025 before, seems our old model didn`t performance well under that ratio. therefore I change the ratio to 0.2 0.1 0.7 in order to beat them.}
\label{tab:my_label2}
\centering
        \begin{tabular}{ c c c c c c c c }
        \hline
         \textbf{Method} & \textbf{Cora} & \textbf{CiteSeer} & \textbf{PubMed} & \textbf{Computers} & \textbf{Photos} & \textbf{CS} & \textbf{Physics}\\
        \hline
         MLP & $66.04 \pm 1.11$ & $68.99 \pm 0.48$ & $82.03 \pm 0.24$& $71.89 \pm 5.36$ & $86.11 \pm 1.35$ & $93.50 \pm 0.24$ &$94.56 \pm 0.11$\\ 
         GCN & $84.72 \pm 0.38$ & $75.04 \pm 1.46$ & $83.19 \pm 0.13$ & $78.82 \pm 1.87$ & $90.00 \pm 1.49$ & $93.00 \pm 0.12$ & $95.55 \pm 0.09$\\ 
         SGC & $83.79 \pm 0.37$ & $73.52 \pm 0.89$ & $75.92 \pm 0.26$ & $77.56 \pm 0.88$ & $86.44 \pm 0.35$ & OOM & OOM\\ 
         GAT & $84.37 \pm 1.13$ & $74.80 \pm 1.00$ & $83.92 \pm 0.28$ &$78.68 \pm 2.09$ & $89.63 \pm 1.75$ & $92.57 \pm 0.14$ & $95.13 \pm 0.15$\\ 
         JKNet & $83.69 \pm 0.71$ & $74.49 \pm 0.74$ & $82.59 \pm 0.54$ & $69.32 \pm 3.94$ & $86.12 \pm 1.12$ & $91.11 \pm 0.22$ & $94.45 \pm 0.33$\\              
         APPNP &  $83.69 \pm 0.71$ &  $75.84 \pm 0.64$ & $80.42 \pm 0.29$ & $73.73 \pm 2.49$ & $87.03 \pm 0.95$ & $91.52 \pm 0.14$ & $94.71 \pm 0.11$\\ 
         GPRGNN & $83.79 \pm 0.93$ &  $75.94 \pm 0.65$ & $82.32 \pm 0.25$ &$74.26 \pm 2.94$ & $88.69 \pm 1.32$ & $91.89 \pm 0.08$ & $94.85 \pm 0.23$\\ 
         UFG & $80.64	\pm	0.74$ & $73.30	\pm	0.19$ & $81.52	\pm	0.80$ & $66.39	\pm	6.09
$ & $86.60	\pm	4.69$ & $95.27	\pm	0.04$ & $95.77	\pm	0.04$\\
         PGNN$^{1.0} $& $84.21 \pm 0.91$ & $75.38 \pm 0.82$ & $84.34 \pm 0.33$ & $81.22 \pm 2.62$ & $87.64 \pm 5.05$ &  $94.88 \pm 0.12$  & $96.15 \pm 0.12$\\  
         PGNN$^{1.5} $& $84.42 \pm 0.71$ & $75.44 \pm 0.98$ & $84.48 \pm 0.21$  & $82.68 \pm 1.15$ & $91.83 \pm 0.77$ & $94.13 \pm 0.08$ & $96.14 \pm 0.08
$\\  
         PGNN$^{2.0} $& $84.74 \pm 0.67$ & $75.62 \pm 1.07$ &$84.25 \pm 0.35$ & $83.40 \pm 0.68$ & $91.71 \pm 0.93$ & $94.28 \pm 0.10$ & $96.03 \pm 0.07$\\
         PGNN$^{2.5} $& $84.48 \pm 0.77$ &$75.22 \pm 0.73$ &$83.94 \pm 0.47$ & $82.91 \pm 1.34$ & $91.41 \pm 0.66$ & $93.40 \pm 0.07$ & $95.75 \pm 0.05$\\
\hline 
         %PGNN*$^{1.0} $& $77.59\pm0.69$ & $63.19\pm0.98$ & $83.21\pm0.30$ & $84.46\pm0.89$ & $90.69\pm0.66$ & $91.46\pm0.50$ & $94.72\pm0.37$\\  
         %PGNN*$^{1.5} $& $78.86\pm0.75$ & $63.80\pm0.79$ & $83.65\pm0.17$ & $85.03\pm0.90$ & $90.91\pm0.50$ & $92.12\pm0.40$ & $94.90\pm0.16$  \\  
         %PGNN*$^{2.0} $& $78.93\pm0.60$ &  $63.65\pm1.08$ & $84.19\pm0.22$ & $84.39\pm0.85$ & $90.40\pm0.63$ & $92.28\pm0.47$ & $94.93\pm0.14$ \\
         %PGNN*$^{2.5} $& $78.87\pm0.57$ & $63.28\pm0.97$ & $84.45\pm0.18$ & $83.85\pm0.87$ & $89.82\pm0.64$ & $91.94\pm0.40$ & $94.87\pm0.11$\\
         
         %\hline
         pL-UFG1$^{1.0}$ & $84.54\pm0.62$ & $75.88\pm0.60$ & $85.56 \pm 0.18$ & $82.07	\pm	2.78$ & $85.57\pm19.92$ &\textcolor{red}{$\mathbf{95.03\!\pm\!	0.22}$} & {$\mathbf{96.19 \!\pm\!0.06}$}\\

         pL-UFG1$^{1.5}$ & $84.96 \pm0.38$ & $76.04 \pm0.85$ & \textcolor{blue}{$\mathbf{85.59\!\pm\!0.18}$}& $85.04 \pm	1.06$ & \textcolor{blue}{$\mathbf{92.92\!\pm\!0.37}$}& \textcolor{red}{$\mathbf{95.03	\pm	0.22}$} & \textcolor{red}{$\mathbf{96.27 \!\pm\!0.06}$}\\
         pL-UFG1$^{2.0}$ & $85.20\pm0.42$ & \textcolor{red}{$\mathbf{76.12\pm0.82}$} & \textcolor{blue}{$\mathbf{85.59\!\pm\! 0.17}$} & $\mathbf{85.26	\!\pm\!1.15}$ & $\mathbf{92.65\!\pm\!0.65}$ &  $94.77	\pm	0.27$ & $96.04 \pm 0.07$\\
         pL-UFG1$^{2.5}$ &  $85.30\pm0.60$ &\textcolor{blue}{$\mathbf{76.11\!\pm\!0.82}$} & $85.54\pm0.18$ & $85.18	\pm	0.88
$ & $91.49\pm1.29$& $94.86	\pm0.14$ & $95.96 \pm 0.11$\\

        \hline
         
         pL-UFG2$^{1.0}$ & $84.42\pm0.32$ &  $74.79 \pm 0.62$ & $85.45\pm0.18$& $84.88	\pm	0.84$ & $85.30\pm19.50$ &  \textcolor{red}{$\mathbf{95.03 \!\pm\!0.19}$}& $96.06\pm0.11$\\
         pL-UFG2$^{1.5}$ & \textcolor{red}{$\mathbf{85.60\!\pm\!0.36}$} & $75.61 \pm0.60$ & \textcolor{blue}{$\mathbf{85.59\!\pm\!0.18}$} & $84.55\pm	1.57$ & \textcolor{red}{$\mathbf{93.00\!\pm\!0.61}$}& \textcolor{red}{$\mathbf{95.03\!\pm\!	0.19}$} & $96.14 \pm 0.09$\\
         pL-UFG2$^{2.0}$ & $85.20\pm0.42$ & \textcolor{red}{\textbf{76.12\(\pm\)0.82}} & \textcolor{blue}{$\mathbf{85.59\!\pm\!0.17}$} & \textcolor{blue}{$\mathbf{85.27\!\pm\!	1.15}$} & $92.50 \pm0.40$ & $94.77\pm	0.27$ & $96.05 \pm 0.07$\\
         pL-UFG2$^{2.5}$ & $\mathbf{85.31\!\pm\!0.41}$ & $75.88\pm0.67$ & $85.53\pm0.19$ & $85.11	\pm	0.81$& $92.44\pm1.51$& $\mathbf{94.96\!\pm\!0.13}$ & $95.99 \pm 0.12$\\
         \hline
         pL-fUFG$^{1.0}$ & $84.45 \pm0.43$ & $75.64 \pm 0.64$ & \textcolor{red}{$\mathbf{85.62 \!\pm\! 0.19}$} & $84.57 \pm 1.08$ & $86.09\pm16.35$ & \textcolor{blue}{$\mathbf{94.98\!\pm\!	0.23}$} &  \textcolor{blue}{$\mathbf{96.23\!\pm\!0.09}$}\\
         pL-fUFG$^{1.5}$ & \textcolor{blue}{$\mathbf{85.40\!\pm\!0.45}$} & $75.94\pm0.63$ & \textcolor{red}{$\mathbf{85.62\!\pm\!0.17}$} & $85.00\pm1.21$ & $92.12\pm1.35$ & \textcolor{blue}{\textbf{94.98\(\pm\)0.23}} & $96.12\pm0.08$\\
         pL-fUFG$^{2.0}$ & $85.20 \pm0.42$ & \textcolor{red}{$\mathbf{76.12\!\pm\!0.82}$ }& $\mathbf{85.58\!\pm\!0.16}$ & \textcolor{red}{$\mathbf{85.29\!\pm\!	1.16}$} & $92.47\pm0.48$ & $94.77	\pm	0.27$ & $96.05 \pm 0.07$\\
         pL-fUFG$^{2.5}$ & $85.21\pm0.44$ & $76.01 \pm0.97$  & $85.54 \pm 0.20$ & $84.94 \pm 0.91$ & $92.26\pm1.26$ & $94.95 \pm	0.11$ & $96.02 \pm 0.04$ \\
        \hline
        \end{tabular}
\end{table*}

\begin{table*}[t]
\setlength{\tabcolsep}{2pt}
\renewcommand{\arraystretch}{1.5}
\caption{Test accuracy (\%) on heterophilic directed graph}
\label{tab:my_label4}
\centering
        \begin{tabular}{ c c c c c c c c }
        \hline
         \textbf{Method} & \textbf{Chameleon} & \textbf{Squirrel} & \textbf{Actor} & \textbf{Wisconsin} & \textbf{Texas} & \textbf{Cornell}\\ 
        \hline
          MLP & $48.82 \pm 1.43$ & $34.30 \pm 1.13$ & \textcolor{red}{$\mathbf{41.66\!\pm\!0.83}$ }& $93.45 \pm 2.09$ & $71.25	\pm	12.99$ & $83.33	\pm	4.55$\\
          GCN & $33.71 \pm 2.27$ & $26.19 \pm 1.34$ & $33.46 \pm 1.42$ & $67.90 \pm 8.16$ &$53.44	\pm	11.23$ & $55.68	\pm	10.57$\\
          SGC & $33.83 \pm 1.69$ & $26.89 \pm 0.98$ & $32.08 \pm 2.22$ &$59.56 \pm 11.19$ &$64.38	\pm	7.53$ &$43.18	\pm	16.41$\\
          GAT & $41.95 \pm 2.65$ & $25.66 \pm 1.72$ & $33.64 \pm 3.45$ &$60.65	\pm	11.08$ & $50.63	\pm	28.36$ & $34.09	\pm	29.15$\\
          JKNet & $33.50 \pm 3.46$ & $26.95 \pm 1.29$ & $31.14 \pm 3.63$ & $60.42	\pm	8.70$ & $63.75	\pm	5.38$ & $45.45	\pm	9.99$\\
          APPNP & $34.61 \pm 3.15$ & $32.61	\pm	0.93$ & $39.11	\pm	1.11$ & $82.41	\pm	2.17$ & $80.00	\pm	5.38$ & $60.98	\pm	13.44$\\
          GPRGNN& $34.23 \pm 4.09$ & $34.01	\pm	0.82$ & $34.63	\pm	0.58$ &$86.11	\pm	1.31$ &$84.38	\pm	11.20$& $66.29	\pm	11.20$\\
        UFG & $50.11	\pm	1.67$ & $31.48	\pm	2.05$ & $40.13	\pm	1.11$ & $93.52	\pm	2.36$ & $84.69	\pm	4.87$ & $83.71	\pm	3.28$\\
          PGNN$^{1.0}$ & $49.04 \pm 1.16$ & $34.79 \pm 1.01$ & $40.91 \pm 1.41$ & $94.35 \pm 2.16$& $82.00 \pm 11.31$ & $82.73 \pm 6.92$ \\
          PGNN$^{1.5}$ & $ 49.12 \pm 1.14$ & $34.86 \pm 1.25$ & $40.87 \pm 1.47$ & $94.72 \pm 1.91$ & $81.50 \pm 10.70$ & $81.97 \pm 10.16$ \\
          PGNN$^{2.0}$ & $49.34 \pm 1.15$ & $34.97 \pm 1.41$ & $40.83 \pm 1.81$ & $94.44 \pm 1.75$ & $84.38 \pm 11.52$ &$81.06 \pm 10.18$ \\
          PGNN$^{2.5}$ & $49.16 \pm 1.40$  
          & $34.94 \pm 1.57$ 
          & $40.78 \pm 1.51$
          & $94.35 \pm 2.16$ 
          & $83.38 \pm 12.95$ 
          & $81.82 \pm 8.86$ \\

          \hline
           %PGNN*$^{1.0}$ & ($48.86\pm1.95$)&($33.75\pm1.50$)&($40.62\pm1.25$)&($95.37\pm2.06$)&($84.06\pm7.41$)&($82.16\pm8.62$)\\
           %PGNN*$^{1.5}$ & ($48.74\pm1.62$)&($33.33\pm1.45$)&($40.35\pm1.35$)&($95.24\pm2.01$)&($84.46\pm7.79$)&($78.47\pm6.87$)\\
           %PGNN*$^{2.0}$ & ($48.77\pm1.87$)&($33.60\pm1.47$)&($40.07\pm1.17$)&($91.15\pm2.76$)&($87.96\pm6.27$)&($72.04\pm8.22$)\\
        %PGNN*$^{2.5}$ & ($48.80 \pm 1.77$)&($33.79 \pm 1.45$)&($39.80 \pm 1.31$)&($87.08 \pm2.69$)&($83.01 \pm6.80$)&($70.31 \pm8.84$)\\
        %\hline
          pL-UFG1$^{1.0}$ & $\mathbf{56.81	\pm	1.69}$ &$38.81\pm	1.97$ & \textcolor{blue}{$\mathbf{41.26\pm1.66}$} & \textcolor{blue}{$\mathbf{96.48 \pm 0.94}$}& $86.13\pm7.47$ & $86.06\pm	3.16$\\
          %pL-UFG1$^{1.0+}$ & $55.60\pm0.90$ & $38.01 \pm	1.50$ & $37.76 \pm	12.69$ &$93.98	\pm 2.63$ & $84.88\pm4.16$ & $87.88\pm2.71$\\
          %pL-UFG1$^{1.2+}$& $58.10\pm1.03$ & $39.55\pm	0.21$ &$42.75\pm	1.92$ & $96.06\pm0.77$ & $87.19\pm	4.18$& $88.26\pm2.48$\\
          
          pL-UFG1$^{1.5}$ &\textcolor{blue}{$\mathbf{56.89\pm1.17}$} & $\mathbf{39.73\pm1.22}$& $40.95\pm0.93$& \textcolor{blue}{$\mathbf{96.48 \pm 1.07}$} & $87.00\pm5.16$ & $\mathbf{86.52\pm	2.29}$\\
          pL-UFG1$^{2.0}$ & $56.24	\pm	1.02$ & $39.72\pm1.86$& $40.95\pm0.93$ & \textcolor{red}{$\mathbf{96.59 \pm 0.72}$} &$86.50\pm8.84$ & $85.30\pm	2.35$\\
          pL-UFG1$^{2.5}$ & $56.11	\pm	1.25$ & $39.38\pm1.78$ & $41.04\pm0.99$ & $95.34 \pm 1.64$ & \textcolor{red}{$\mathbf{89.00\pm4.99}$}& $83.94\pm3.53$\\
          \hline
          pL-UFG2$^{1.0}$ & $55.51	\pm	1.53$ &$36.94\pm5.69$ & $29.28\pm19.25
$& $93.98 \pm 2.94$ & $85.00 \pm 5.27$ & \textcolor{red}{$\mathbf{87.73 \pm 2.49}$} \\
         %pL-UFG2$^{1.0+}$ & $54.70\pm1.79$ & $36.13\pm5.34$ & $32.69 \pm	16.65$  & $93.98	\pm 2.63$ & $83.50\pm3.25$ & $88.48	\pm3.33$\\ 
         %pL-UFG2$^{1.2+}$ & $57.99\pm0.51$ & $39.05\pm0.74$ & $38.37\pm	5.048$& $93.32\pm3.95$& $86.56\pm	4.09$ &$87.12\pm	2.73$\\

         pL-UFG2$^{1.5}$ &  \textcolor{red}{$\mathbf{57.22 \pm 1.19}$} & \textcolor{red}{$\mathbf{39.80 \pm 1.42}$} & $40.89\pm 0.75
$ & \textcolor{blue}{$\mathbf{96.48 \pm 0.94}$}& $87.63 \pm 5.32$ & $86.82\pm	1.67$\\
          pL-UFG2$^{2.0}$ & $56.19	\pm	0.99$ & \textcolor{blue}{$\mathbf{39.74\pm1.66}$} & $41.01\pm0.80$ & $96.14 \pm 1.16$& $86.50\pm	8.84$ & $85.30\pm	2.35$\\
          pL-UFG2$^{2.5}$ & $55.69	\pm	1.15$ & $39.30\pm1.68$& $40.86\pm0.74$ & $95.80 \pm 1.44$& $86.38\pm	2.98$ & $84.55 \pm 3.31$\\
          \hline
          pL-fUFG$^{1.0}$  & $55.80\pm1.93$ & $38.43\pm1.26$ & $32.84\pm16.54$ & $93.98 \pm 3.47$ & $86.25 \pm 6.89$ & \textcolor{blue}{$\mathbf{87.27 \pm 2.27}$}\\
          %pL-fUFG$^{1.0+}$ & $54.31\pm1.92$ & $38.24\pm4.12$ & $36.61 \pm	12.32$ & $93.33\pm2.95$ & $82.63\pm6.11$ & $87.88\pm3.25$\\
          %pL-fUFG$^{1.2+}$ & $58.21\pm0.89$ & $38.78\pm	0.53$ & $40.64\pm2.71$ & $95.14\pm1.20$ & $85.94\pm	2.71$ &$86.74\pm	2.91$\\
          pL-fUFG$^{1.5}$& $55.65	\pm	1.96$ & $38.40\pm1.52$& $41.00\pm0.99$ & \textcolor{blue}{$\mathbf{96.48 \pm 1.29}$}& $87.25\pm3.61$ & $86.21\pm	2.19$\\
          pL-fUFG$^{2.0}$ & $55.95	\pm	1.29$ &$38.33\pm1.71$ & $\mathbf{41.25\pm0.84}$ & $\mathbf{96.25 \pm 1.25}$ & \textcolor{blue}{$\mathbf{88.75\pm4.97}$} & $83.94\pm	3.78$\\
          pL-fUFG$^{2.5}$ & $55.56	\pm	1.66$ & $38.39\pm1.48$& $40.55\pm0.50
$ & $95.28\pm2.24$& $\mathbf{88.50\pm7.37}$ & $83.64 \pm 3.88$\\
        \hline
        \end{tabular}
\end{table*}

\subsection{Experiments on Denoising Capacity} \label{ssec: num3}
\begin{figure}[t]
    \centering
    \includegraphics[scale=0.3]{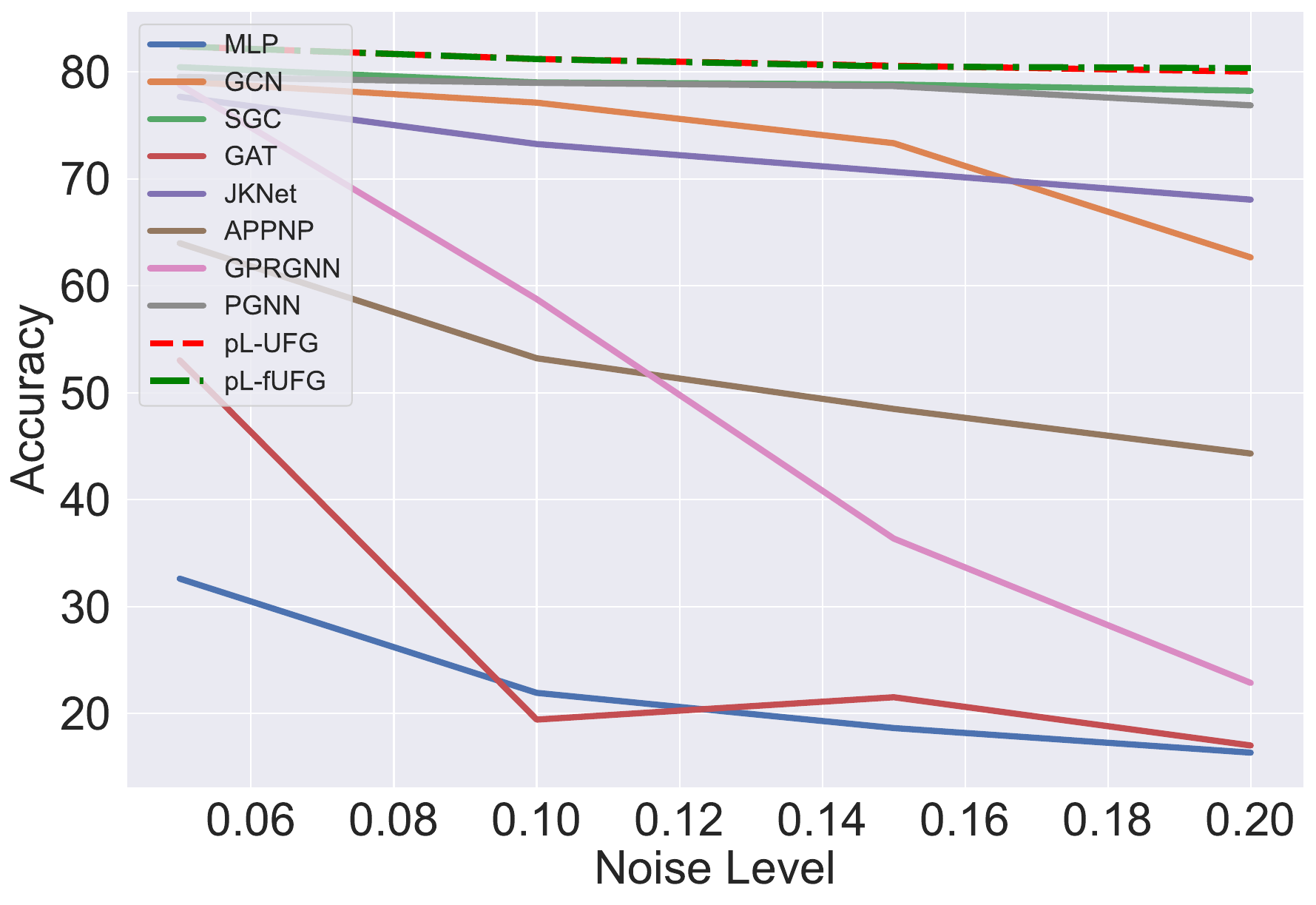}
    \caption{Denoising power on homophilic graph (Cora)} 
    \label{fig:denoise_cora}
\end{figure}

\begin{figure}[t]
    \centering
    \includegraphics[scale=0.3]{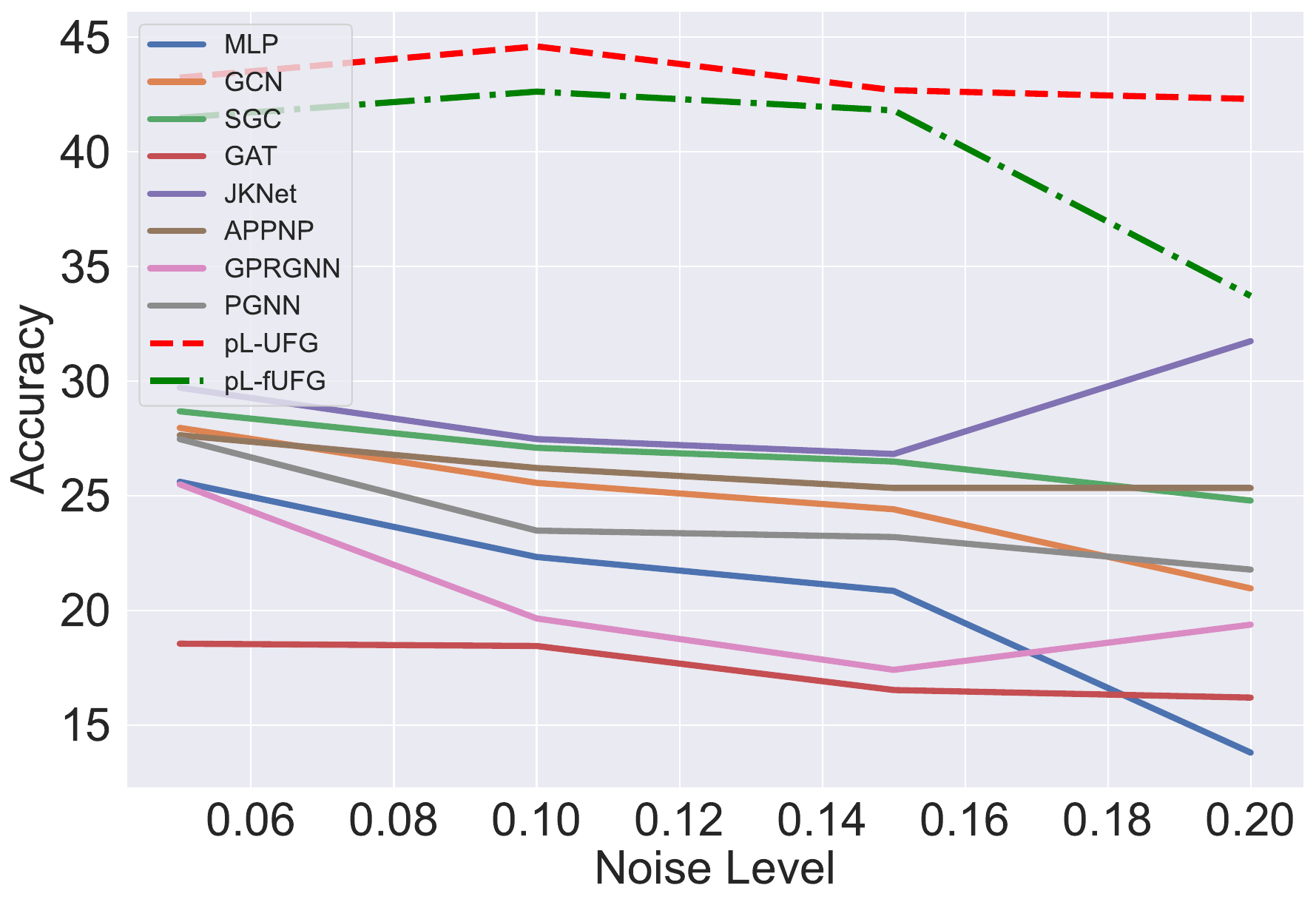}
    \caption{Denoising power on heterophilic graph (Chameleon)} 
    \label{fig:denoise_chamel}
\end{figure}

Followed by the discussion in Section \ref{discussion_denoising_power}, \EScomment{in this section we evaluate our proposed models' (pL-UFG and pL-fUFG) denoising capacity on both homophilic (\texttt{Cora}) and heterophilic (\texttt{Chameleon}) graphs. }Since the node features of the included datasets are in binary, we randomly assign binary noise with
% we randomly assign binary noise \JScomment{since the feature of datasets are in binary,} we randomly assign binary noise \JScomment{with uniform distribution on graph features.}\question{Please further specify on why only binary noise is added here} to 
different proportions of the node features (i.e., $r \in \{5\%, 10\%, 15\%, 20\%\}$). From Fig. \ref{fig:denoise_cora} and \ref{fig:denoise_chamel} we see that pL-UFG$2^{1.5}$ defined in Eq. \eqref{Eq:1variant} outperforms other baselines and showed the strongest robustness to the contaminated datasets. This is expected based on our discussion on the denoising power in Section \ref{discussion_denoising_power} and the formulation of our models (defined in Eqs. \eqref{Eq:1}, \eqref{Eq:1variant} and \eqref{Eq:4}). The denoising capacity of our proposed model is sourced from two parts including the sparse approximation of the framelet decomposition and reconstruction, i.e., $\mathcal W$ and $\mathcal W^T$, and variational regularizer $\mathcal S_p(\mathbf F)$ \citep{Dong2017}. Compared to pL-UFG1 defined in Eq. \eqref{Eq:1}, pL-UFG2 assigns  the denoising operators, $\mathcal W$, $\mathcal W^T$ and $\mathcal S_{p}(\mathbf F)$, to the node inputs that is reconstructed from both sparsely decomposed low and high-frequency domains so that the denoising operators target on the original graph inputs at different scales and hence naturally result in a better model denoising performance. In addition, without taking the reconstruction in the first place, the term in pL-fUFG $\| \mathbf{F}_{r,j} - \textrm{diag}(\theta_{r,j}) \mathcal{W}_{r,j}  \mathbf X \|^2_F$ is less sparse than the pL-UFG2 counterpart. Thus regularizing with the same $\mathcal S_p(\mathbf F)$, the effect from insufficient eliminated noise from pL-fUFG will lead pL-fUFG to an incorrect (respect to pL-UFG2) solution space which can not be recovered by the additional reconstruction using $\mathcal W^T$, and this is the potential reason for observing an outperforming result from pL-UFG2 compared to other proposed models.    

\EScomment{
In addition to the above experiment, here we show how the results of how the quantity of $p$ affect the denoising power of the proposed model, while keeping all other parameters constant. The results of the changes of the denoising power are included in Fig.~\ref{denosing_homo} (homophilic graph) and \ref{denosing_heter} (heterophilic graph).

\begin{figure}[t]
    \centering
    \includegraphics[scale = 0.4]{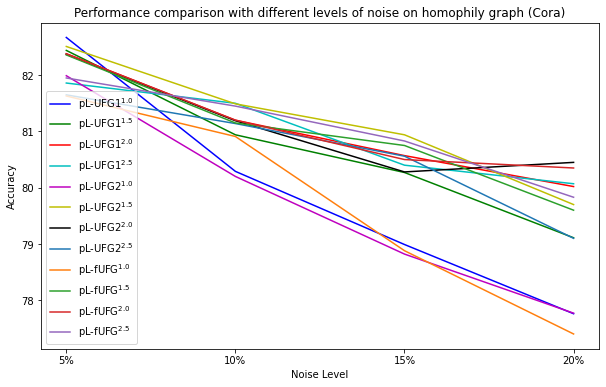}
    % \caption{The denoising power by comparing with different $p$ in homophilic graph}
    \caption{Changes of the denoising power by the quantity of $p$ via homophilic graph.}
    \label{denosing_homo}
\end{figure}

\begin{figure}[t]
    \centering
    \includegraphics[scale = 0.4]{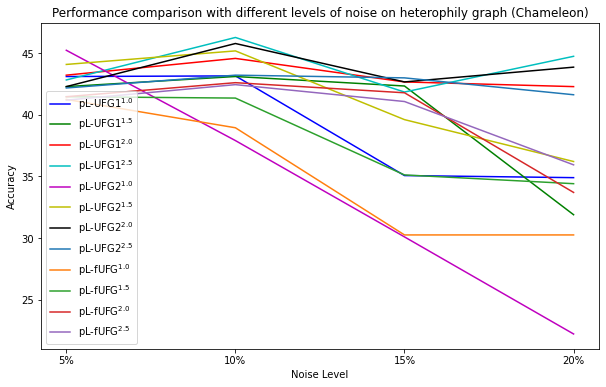}
    \caption{Changes of the denoising power by the quantity of $p$ via heterophilic graph.}
    \label{denosing_heter}
\end{figure}
Based on the observations from the results, it appears that the performance of different values of $p$ varies under different noise rates. For the homophilic graph, the performance of pL-UFG2$^{1.5}$ and pL-UFG2$^{2.0}$ seems to be relatively stable across different noise rates, while the performance of other values of $p$ fluctuates more. For the heterophilic graph, the performance of pL-UFG1$^{2.5}$ and pL-UFG2$^{2.5}$ appears to be relatively stable across different noise rates, while the performance of other values of $p$ fluctuates more.

We also note that these observations on the fluctuations of the results are expected, as we have shown that the denoising process of our proposed model can be presented as: 
\begin{equation}
\mathbf F \!=\! \argmin_{\mathbf{F}} \!\sum_{(r,\ell)\in\mathcal Z} \|\nabla \mathbf F_{(r,\ell)}\|_p^p +\!\mu \|\mathbf{F} \!-\! \mathcal{W}^T\!\textrm{diag}(\!\theta\!)\mathcal{W}\mathbf X\|^2_F. 
\end{equation} 
where $\mathcal Z = \{ (r,\ell) : r = 1,...,R, \ell= 1,...,J \} \cup \{ (0,J)\}$ is the set of indices of all framelet decomposed frequency domains. It is not hard to verify that once a larger quantity of $p$ is assigned, the penalty on the node feature difference ($\nabla \mathbf F$) becomes to be greater. Therefore, a stronger denoising power is induced. In terms of different graph types, when the input graph is heterophily, in most of the cases, the connected nodes tend to have distinct features, after assigning noise to those features, if the feature difference becomes larger, a higher quantity of $p$ is preferred. In the meanwhile, adding noise for the heterophilic graph could also make the feature difference becomes smaller, in this case a large quantity of $p$ may not be appropriate, this explains why most of lines in Fig.~\ref{denosing_heter} are with large fluctuations. Similar reasoning can be applied to the case of the denoising performance of our model for homophilic graphs, we omit it here.
}

\EScomment{
\subsection{Regarding to Ablation Study} 
It is worth noting that one of the advantages of assigning an adjustable $p$-Laplacian-based regularizer is on the convenience of conducting the ablation study. As the key principle of ablation study is to test whether a new component (in our case the regularizer) in a proposed method can always add advantages over
baseline model counterparts regardless of the newly involved parameters. This suggests that the ablation study of our method are naturally conducted to compare our models with the regularizers with the underlying networks  via both the node classification tasks and the model denoising power test. It is not hard to observed that in most of cases, regardless of the change of $p$, our proposed model kept outperforming baseline models. However, as our proposed model was driven from an regularization framework, another potential parameter that we note could affect the model performance is the quantity of $\mu$. Based on Eq.~\eqref{Eq:Model3}, an increase of the quantity of $\mu$ leads to a increase of model's convexity, as a consequence, could guide model more closed to the global optima when $p \neq 2$. Another observation to $\mu$ is we found that a smaller value of $\mu$ together with a relatively bigger value of $p$ are more suitable for homophily/heterophilic graphs, this implies that $\mu$ seems to have an opposite effect on model's adaption power compared to $p$. However, to quantify the effect of $\mu$ via a suitable measure (i.e., potentially Dirichlet energy \citep{han2022generalizedenergy}) is out of the scope of this paper, we leave it to future discussions.

}

%\subsection{Experiments on Oversmoothing tolerance}

%%%%%%%%%%%%%%%%%%%%%%%%%%%%%%%%%%%%%%%%%%
%%%%%%%%%%%%%%%   Notation  %%%%%%%%%%%%%%
%%%%%%%%%%%%%%%%%%%%%%%%%%%%%%%%%%%%%%%%%%

\section{Conclusion and Further work}\label{Sec:6}
\EScomment{
This paper showcases the application of $p$-Laplacian Graph Neural Networks (GNN) in conjunction with framelet graphs. The incorporation of $p$-Laplacian regularization brings remarkable flexibility, enabling effective adaptation to both homophilic undirected and heterophilic directed graphs, thereby significantly enhancing the predictive capabilities of the model. To validate the efficacy of our proposed model, we conducted extensive numerical experiments on diverse graph datasets, demonstrating its superiority over baseline methods. Notably, our model exhibits robustness against noise perturbation, even under high noise levels. These promising findings highlight the tremendous potential of our approach and warrant further investigations in several directions. For instance, delving into the intriguing mathematical properties of our model, including weak and strong convergence, analyzing the behavior of (Dirichlet) energy, and establishing connections with non-linear diffusion equations, opens up fascinating avenues for future research. } 

% \JScomment{This paper introduces a novel framelet graph approach based on $p$-Laplacian GNN. The $p$-Laplacian regularization provides great flexibility to make framelet adaptive to both homophilic undirected and heterophilic directed graphs and therefore largely enhances the model's predictive power. Extensive numerical experiments demonstrate that our proposed model excels against benchmarks on a variety of graph datasets, affirming its robustness against noise perturbation, even at high noise levels. The positive results show great potential and encourage further exploration or applications to other real-world tasks such as image processing and computer vision applications, as it aids in smoothing out noisy or jagged images while still maintaining crucial features, the related studies such as graph generation \citep{ho2020denoising} and graph-based time series forecasting \citep{wen2023diffstg}, with increased flexibility. The positive results show great potential and encourage further exploration. Our future study includes developing an adaptive $p$-Laplacian regularization scheme separately on both low and high-frequency spectral domains.
% }

\bibliographystyle{spbasic}
\bibliography{sn-bibliography}% common bib file


\end{document}